\documentclass[lettersize,journal]{IEEEtran}
\usepackage{amsmath,amsfonts}
\usepackage{array}
\usepackage[caption=false,font=normalsize,labelfont=sf,textfont=sf]{subfig}
\usepackage{textcomp}
\usepackage{stfloats}
\usepackage{url}
\usepackage{verbatim}
\usepackage{graphicx}
\def\BibTeX{{\rm B\kern-.05em{\sc i\kern-.025em b}\kern-.08em
    T\kern-.1667em\lower.7ex\hbox{E}\kern-.125emX}}
\usepackage{balance}

\usepackage{newfloat}
\usepackage{listings}
\usepackage{cite}

\usepackage{algorithm}
\usepackage{algpseudocode}
\usepackage{amsmath}
\usepackage{amssymb}

\usepackage{booktabs}
\usepackage{multirow}
\usepackage{multicol}
\usepackage{graphicx}
\usepackage{xcolor}
\usepackage[table]{xcolor}
\usepackage{pifont}     
\usepackage{amssymb}    
\usepackage{amsmath}    
\usepackage{adjustbox}
\usepackage{makecell}

\usepackage{hyperref}
\hypersetup{
    colorlinks=true,
    linkcolor=blue,
    filecolor=magenta,      
    urlcolor=cyan,
    citecolor=blue,
}

\definecolor{new}{RGB}{133,133,156}

\usepackage{tikz}
\newcommand{\gradientstar}{\includegraphics[height=1em]{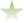}}
\newcommand{\bluestar}{\includegraphics[height=1em]{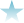}}
\newcommand{\greensquare}{\includegraphics[height=1em]{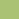}}
\newcommand{\bluesquare}{\includegraphics[height=1em]{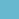}}
\newcommand{\graytriangle}{\includegraphics[height=1em]{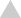}}

\begin{document}
\title{AV-Master: Dual-Path Comprehensive Perception Makes Better Audio-Visual Question Answering}
\author{Jiayu Zhang, Shuo Ye, Qilang Ye, Xun Lin, Zihan Song, Zitong Yu, \IEEEmembership{Senior Member, IEEE}
\thanks{This work was supported by National Natural Science Foundation of China (Grant No. 62306061 and 62576076), Guangdong Research Team for Communication and Sensing Integrated with Intelligent Computing (Project No. 2024KCXTD047), and sponsored by CCF-Tencent Rhino-Bird Open Research Fund. Jiayu Zhang and Shuo Ye contributed equally to this work. Corresponding author: Zitong Yu.

Jiayu Zhang, Zihan Song and Zitong Yu are with Great Bay University, and Dongguan Key Laboratory for Intelligence and Information Technology.

Qilang Ye is with Nankai University. 

Shuo Ye is with Great Bay University. He is also with the Tsinghua Shenzhen International Graduate School, Tsinghua University

Xun Lin is with The Chinese University of Hong Kong.

}}

\markboth{}%
{How to Use the IEEEtran \LaTeX \ Templates}

\maketitle


\begin{abstract}
Audio-Visual Question Answering (AVQA) requires models to effectively utilize both visual and auditory modalities to answer complex and diverse questions about audio-visual scenes. However, existing AVQA methods still face two fundamental challenges: accurately locating question-relevant temporal evidence in long audio-visual streams and adaptively exploiting the dominant modality when audio and visual cues contribute unequally across different questions. This limits their reasoning capability in complex scenarios. To address these challenges, we propose a novel framework named AV-Master. It enhances the model's ability to extract key information from complex audio-visual scenes with substantial redundant content by dynamically modeling both temporal and modality dimensions. In the temporal dimension, we introduce a dynamic adaptive focus sampling mechanism that adaptively aggregates audio-visual evidence with temporal informativeness, effectively mitigating redundancy and segment fragmentation in traditional sampling methods. In the modality dimension, we propose a preference-aware strategy that models each modality's contribution independently, enabling selective activation of critical features. Furthermore, we introduce a dual-path contrastive loss to enhance the collaborative consistency between the temporal dynamic perception path and the global preference activation path, while simultaneously preserving the complementary audio and visual cues for answer prediction. Experiments on four large-scale benchmarks show that AV-Master significantly outperforms existing methods, especially in complex reasoning tasks. The code is available at: {\hypersetup{urlcolor=black} \url{https://github.com/AoKoo233/AV-Master}}.

\end{abstract}


\section{Introduction}

\begin{figure}[!t]
  \centering
  \includegraphics[width=\linewidth]{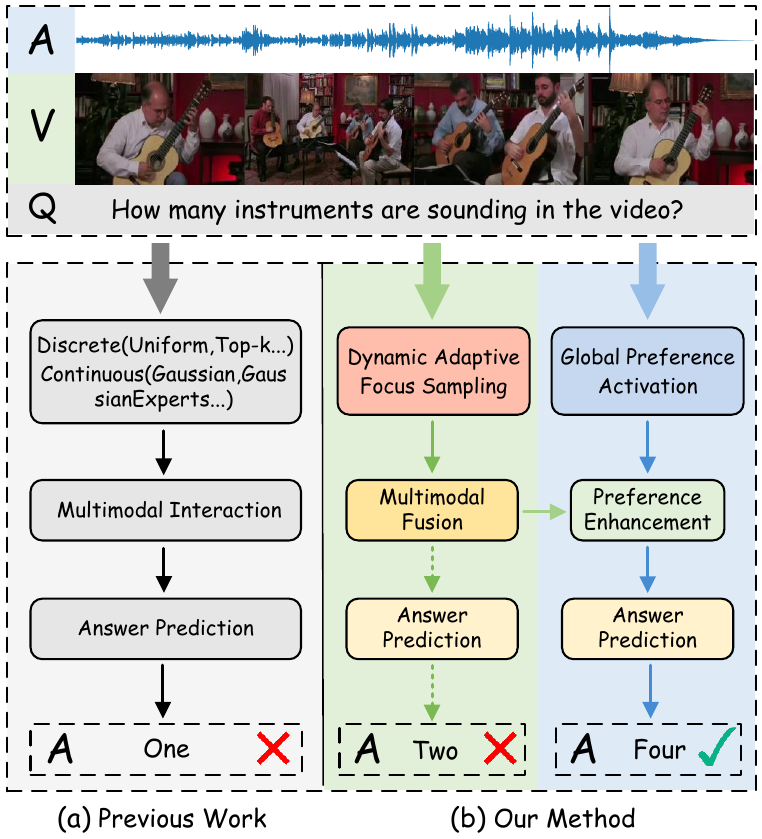}
  \caption{Illustration of the AVQA task and the comparison of our method with previous work. Our method employs dynamic adaptive focus sampling to capture key audio-visual segments and predicts modal preferences through global preference activation to enhance the model, ultimately generating the correct answer.}
  \label{fig:1}
\end{figure}

Humans perceive the world through various modalities, such as vision, hearing, and touch. Inspired by such multisensory experiences, researchers have increasingly focused on a range of multimodal understanding tasks~\cite{zhang2026multimodal,tang2025srvc,lin2025svc,wang2026micro}. Among these tasks, Audio-Visual Question Answering~(AVQA)~\cite{li2023progressive,li2023multi,kim2025question} is a practical task with broad application prospects. AVQA leverages both visual and auditory modalities, requiring models to perform fine-grained, question-conditioned discrimination over short temporal segments and subtle modality-specific cues, rather than coarse understanding of entire video clips as in other tasks (e.g., video captioning~\cite{zhang2025mgtr}). For instance, to answer a question such as ``Which flute produces sound first?'', the model must identify a specific brief moment and distinguish subtle cross-modal differences among similar candidate objects. This process involves dynamically understanding audio-visual segments and addressing question-specific modality preferences, significantly increasing the complexity of the task. 


For AVQA, we argue it is essential to focus on the following: \textit{(i) How can the model identify the most relevant visual and audio segments of the video related to the given question? (ii) How can the model appropriately model the question-conditioned modality dependency, thereby flexibly utilizing audio-specific, visual-specific, and joint multimodal evidence for answer prediction?} For (i), most current models employ discrete~\cite{li2022learning,li2024boosting} or continuous~\cite{kim2025question} sampling methods. Although these methods have achieved promising results, they still have notable limitations. The former compromises the inherent temporal cues in audio-visual segments, resulting in hallucinations during fine-grained understanding. The latter partially alleviates this issue but introduces significant redundancy, which hinders further enhancement of model performance. For~(ii), most mainstream methods are not specifically designed to differentiate between various input modalities but instead treat them merely as supplementary information. A few studies~\cite{zhao2025audio} have recognized these challenges and design modality-specific fusion weights during multimodal interaction, thereby guiding the model to favor a particular modality. However, such methods typically estimate modality importance only after multimodal interaction has been completed. At this stage, audio and visual evidence are already intertwined, so the resulting weights mainly serve to rebalance the fused features rather than explicitly enhancing modality-specific evidence that is closely related to the given question. As a result, their ability to guide performance improvement is limited.

To solve these issues, we propose AV-Master, which can find the important fine-grained areas related to the current question within complex dynamic audio-visual scenes, and generate the optimal answer by integrating global preference information. As shown in Fig.~\ref{fig:1}, unlike previous work~\cite{li2024boosting,kim2025question}, AV-Master employs dynamic adaptive focus sampling to extract the fine-grained focus features from the audio-visual segments relevant to the current questioning scenario. The focus sampling combines the advantages of discrete~($e.g.$, uniform~\cite{li2022learning}, top-k~\cite{li2024boosting}) and continuous~($e.g.$, gaussian, gaussian-experts~\cite{kim2025question}) sampling methods, allowing the model to capture all continuous time steps while significantly reducing redundant information obtained from sampling. 
Furthermore, we propose a global preference activation strategy focused on model modality preferences. 
This strategy determines the preference distribution of the model in different modalities through independent perception and improves the decision-making capabilities of the model. 
Compared to using dynamic weights at the multimodal fusion stage, our enhancement strategy focuses on the global information in the initial features. This design enables the model to learn audio and visual preferences under question guidance based on their respective unimodal representations, thereby providing complementary global cues to the temporal dynamic perception path that focuses on capturing fine grained temporal evidence.
Our contributions are summarized as follows:

\begin{itemize}
    \item We propose a dual-path audio-visual learning model named AV-Master that enhances the understanding of audio-visual scenes by perceiving fine-grained details and global modality preferences related to questions, achieving cross-modal mapping from audio-visual signals to textual answers.
    \item We introduce a dynamic adaptive focus sampling method that progressively performs adaptive learning from the input audio-visual segments during the encoding, enabling precise capture of the focal areas within a large number of redundant segments.
    \item Extensive evaluation on four benchmark AVQA datasets demonstrates that our proposed AV-Master is superior and achieves new state-of-the-art performance compared to existing AVQA methods.
\end{itemize}


The remainder of this paper is organized as follows. In
Section~\ref{sec:2}, we review the related work relevant to the research
direction of this study. Section~\ref{sec:3} provides a detailed description of the architecture of the proposed model. Section~\ref{sec:4}
presents experimental results that validate the effectiveness of
our proposed methods. Finally, in Section~\ref{sec:5}, we conclude by
summarizing the main contributions of this paper.

\vspace{-0.1em}
\section{Related Work}\label{sec:2}

\subsection{Audio-Visual Scene Understanding}

Audio-visual scene understanding aims to learn semantically consistent and temporally aligned representations from visual and auditory streams. Early studies in this area mainly focused on cross-modal correspondence and low-level perception, such as sound source localization and audio-visual synchronization, where the goal is to associate a sound with its visual origin or to determine whether the two modalities are temporally matched~\cite{risoud2018sound,gan2020music,gao2021visualvoice}. Building on this line of research, subsequent works extended audio-visual learning to more structured perception tasks, including action recognition~\cite{gao2020listen,chen2023agpn,shaikh2024multimodal}, event detection~\cite{tian2018audio,xuan2020cross,zhou2022contrastive}, video parsing~\cite{wu2021exploring,lai2023modality,zhou2024label}, and audio-visual source separation~\cite{gao2018learning,gao2019co,chen2024bootstrapping}. More recent efforts further emphasize robust multimodal representation learning in complex dynamic environments, promoting the transition from simple cross-modal alignment to higher-level scene understanding~\cite{ye2024dep,yu2023pvass,zeng2025exploring}. This evolution shows that audio-visual learning has gradually moved from low-level correspondence modeling toward semantic reasoning over dynamic scenes. However, most existing audio-visual scene understanding tasks are still formulated as recognition, localization, or correspondence problems. In contrast, audio-visual question answering requires the model not only to align audio and visual cues, but also to identify question-relevant evidence, reason over temporally distributed events, and flexibly determine which modality should dominate the final decision. Therefore, AVQA can be viewed as a more challenging extension of audio-visual scene understanding.

\subsection{Audio-Visual Question Answering}

Audio-Visual Question Answering~\cite{jiang2023target,jiang2025clip} (AVQA) extends traditional visual question answering~\cite{10547379}~(VQA) and audio question answering~\cite{li2025reinforcementlearningoutperformssupervised}~(AQA) by requiring joint reasoning over video, audio, and language. Early attempts largely followed the paradigm of multimodal fusion developed for VQA or video reasoning, where models aggregate audio-visual features and answer questions based on global fused representations. Representative methods such as HME~\cite{fan2019heterogeneous}, PSAC~\cite{li2019beyond} and HCRN~\cite{le2020hierarchical} demonstrated the feasibility of multimodal question answering, but their reasoning process was still dominated by relatively coarse feature aggregation and lacked AVQA-specific temporal modeling.

With the release of dedicated benchmarks, research shifted from general multimodal fusion to task-specific spatiotemporal reasoning. ST-AVQA~\cite{li2022learning} explicitly introduced spatiotemporal modeling for music-oriented AVQA, showing that fine-grained temporal and spatial cues are essential for answering audio-visual questions. LAVISH~\cite{lin2023vision} further strengthened audio-visual-language interaction, improving multimodal fusion for more challenging reasoning scenarios. These works established the importance of structured multimodal reasoning, but they still mainly relied on holistic feature encoding and fusion.

More recent AVQA methods have increasingly focused on identifying question-relevant temporal evidence. TSPM~\cite{li2024boosting} improves AVQA by constructing template-based prompts and enhancing temporal/spatial perception, thereby helping the model locate informative segments related to the question. QA-TIGER~\cite{kim2025question} further explores adaptive temporal selection by combining continuous and discontinuous frame modeling, which alleviates part of the redundancy problem in dense temporal reasoning. Along another direction, AVAF-Net~\cite{zhao2025audio} introduces adaptive fusion weights to account for the varying importance of audio and visual modalities under different questions. Recent methods such as SHMamba~\cite{yang2025shmamba}, PSOT~\cite{li2025patch}, and CoQo~\cite{pei2025guiding} continue this trend by enhancing sequence modeling, training strategy, or backbone representation, leading to stronger overall AVQA performance.

Despite this progress, two limitations remain insufficiently addressed. First, existing methods often model temporal evidence either through coarse discrete selection or dense continuous aggregation. The former may destroy temporal continuity and miss subtle yet important cues, while the latter tends to introduce considerable redundancy, making subsequent reasoning more difficult. Second, although some methods consider modality importance during multimodal fusion, modality preference is usually inferred only after the two modalities have been entangled. As a result, the model may still struggle to explicitly capture question-conditioned modality dominance from the original unimodal representations. Different from previous methods, our approach addresses these two issues in a unified manner by combining dynamic adaptive focus sampling for fine-grained temporal perception with a global preference activation strategy for modality-aware reasoning.

\begin{figure*}
    \centering
    \includegraphics[width=\textwidth, clip]{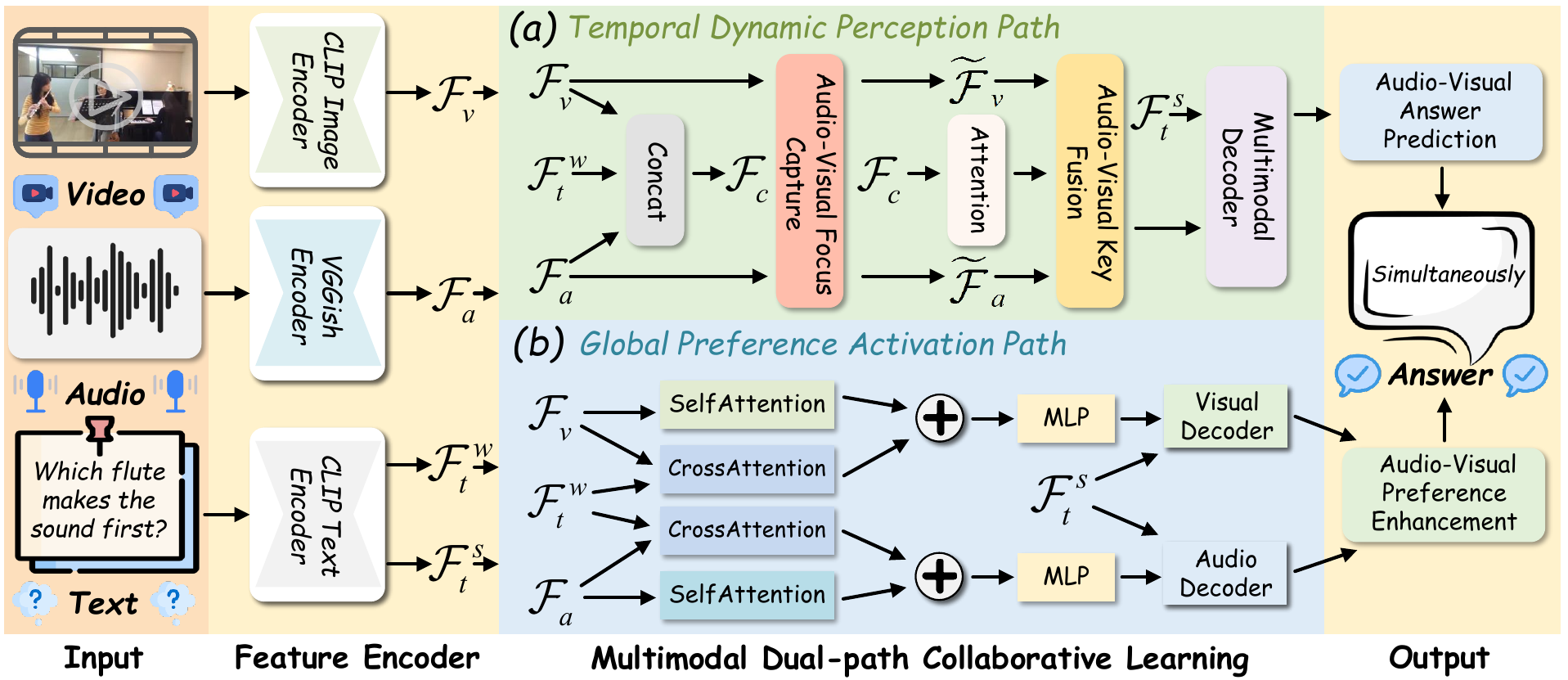}
    \caption{Overview of AV-Master. We utilize three separate pre-trained encoders to extract features from video, audio, and question inputs. The encoded features are then fed into the temporal dynamic perception path and the global preference activation path, respectively. Finally, the model predicts the correct answer based on the outputs of these two paths.}
    \label{fig:2}
\end{figure*}

\section{Methodology}\label{sec:3}

\begin{figure*}
    \centering
    \includegraphics[width=\textwidth, clip]{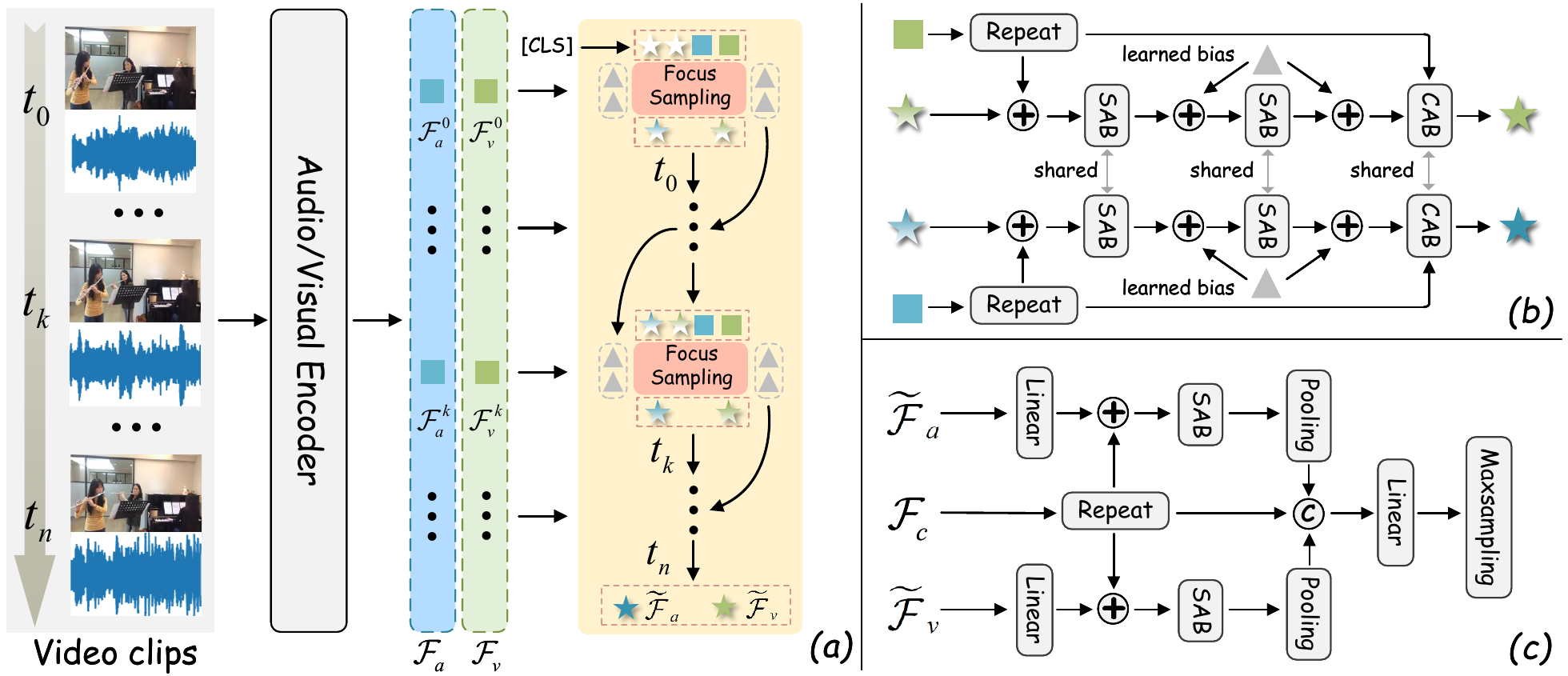}
    \caption{The pipeline of \textit{(a)~audio-visual focus capture} and \textit{(c)~audio-visual key fusion} in the \textbf{temporal dynamic perception path}, where \textit{(b)} represents the specific implementation process of focus sampling in \textit{(a)~audio-visual focus capture}. SAB and CAB represent the self-attention block and the cross-attention block, respectively. \bluestar \hspace{0.5em} \gradientstar \hspace{0.2em} represent the input predefined CLS tokens (serve as audio-visual templates), \bluesquare \hspace{0.5em} \greensquare \hspace{0.2em} represent the audio-visual features at a certain time step, and \graytriangle \hspace{0.2em} is a learned bias.}
    
    \label{fig:3}
\end{figure*}

In this section, we will introduce our proposed AV-Master in detail. Specifically, we first introduce the feature information used and the feature extraction settings in Section~\ref{sec:input-rep}. Subsequently, in Section~\ref{sec:3B} and Section~\ref{sec:3C}, we respectively explain the proposed temporal dynamic perception path and global preference activation path. Finally, in Section~\ref{sec:3D}, we describe in detail the learning objectives involved in the training phase, which include the dual-path prediction loss and the dual-path contrastive loss. The overall architecture is shown in Fig.~\ref{fig:2}, and the specific implementation flow of the temporal dynamic perception path proposed in Fig.~\ref{fig:2} is illustrated in Fig.~\ref{fig:3}.

\subsection{Input Representation}\label{sec:input-rep}

\textit{(a)~Visual representation:} For a given video, we split it into $T$ non-overlapping 1$s$ segments, each with paired audio and visual elements. Each visual segment is processed by a pre-trained vision-language model CLIP~\cite{radford2021learning}. In this process, a special token is added at the beginning of each segment and is used as the visual feature. The visual features can be represented as~$\mathcal{F}_{v}=\left\{\mathcal{F}_{v}^{0},\mathcal{F}_{v}^{1},\cdots,\mathcal{F}_{v}^{T-1}\right\}$ $\in \mathbb{R}^{T \times D}$, where $D$ denotes the feature dimension.

\textit{(b)~Audio representation:} For each audio segment, we follow previous works~\cite{li2024boosting,li2023progressive,kim2025question} that use the pre-trained VGGish model~\cite{hershey2017cnn} to extract audio features. The audio features can be represented as~$\mathcal{F}_{a}=\left\{\mathcal{F}_{a}^{0},\mathcal{F}_{a}^{1},\cdots,\mathcal{F}_{a}^{T-1}\right\}$ $\in \mathbb{R}^{T \times D}$. The parameters of the CLIP and VGGish models are frozen during training.

\textit{(c)~Question representation:} For the input question, we use the CLIP text encoder to obtain word-level features $\mathcal{F}_{t}^{w}=\left\{\mathcal{F}_{t}^{1},\mathcal{F}_{t}^{2},\cdots,\mathcal{F}_{t}^{L}\right\}$ $\in \mathbb{R}^{L \times D}$, and extract the sentence-level feature 
$\mathcal{F}_{t}^{s}$ $\in \mathbb{R}^{1 \times D}$ by taking the embedding of the first token. $L$ denotes the number of question tokens.

\subsection{Temporal Dynamic Perception Path}\label{sec:3B}

To extract key information from complex initial audio-visual features, we propose a temporal dynamic perception path, which consists of two modules: an audio-visual focus capture module and an audio-visual key fusion module. The former focuses more on the relationships within each modality, aiming to discover key hidden clues in the current modality, while the latter concentrates on establishing connections between different modalities. Additionally, the inputs to the perception path are the outputs from the previous feature encoding stage, including visual features $\mathcal{F}_{v}$, audio features $\mathcal{F}_{a}$, word-level features $\mathcal{F}_{t}^{w}$, and sentence-level features $\mathcal{F}_{t}^{s}$.

\textit{(a)~Audio-visual focus capture:} As shown in Fig.~\ref{fig:3}~(a), to achieve fine-grained perception, the visual features $\mathcal{F}_{v}$ and the audio features $\mathcal{F}_{a}$ are fed into the audio-visual focus capture module for dynamic adaptive focus sampling at each time step from time $0$ to time $n$, where $n = T - 1$. During the perception process, we utilize predefined templates~($\mathcal{A}_{cls}$ and $\mathcal{V}_{cls}$) to focus on the audio-visual features at each moment, thereby extracting global effective information from the complex initial features. Moreover, we introduce learnable biases to highlight key regions with significant variations between different moments for improving sampling accuracy. Over time, the predefined templates progressively extract the critical information hidden at different moments throughout the entire audio-visual segment. Finally, we take the templates from the $n$-th time step as the output and obtain the fine-grained focus features $\widetilde{\mathcal{F}}_v$ and $\widetilde{\mathcal{F}}_a$.

\textit{(b)~Dynamic adaptive focus sampling:} For a specific time step $k$ , the detailed focus sampling process is shown in Fig.~\ref{fig:3}~(b). First, the $k$-th time step visual feature $\mathcal{F}_{v}^{k}$ and audio feature $\mathcal{F}_{a}^{k}$ together with the $(k-1)$-th time step template $\mathcal{A}_{cls}^{k-1}$ and $\mathcal{V}_{cls}^{k-1}$ serve as inputs for the entire sampling process. Features $\mathcal{F}_{v}^{k}$ and $\mathcal{F}_{a}^{k}$ are repeated to match the length of their corresponding templates and then added to the templates. The resulting sums are fed into a self-attention block~(SAB) for attention enhancement. These enhanced features are then added to a learnable bias and go through the above operation once more. Afterward, these features, along with the repeated feature, are passed into a cross-attention block~(CAB) for modality interaction. Finally, we can obtain the updated templates $\mathcal{A}_{cls}^{k}$ and $\mathcal{V}_{cls}^{k}$ for the current time step. This process is formulated as follows:

\begin{equation}
    \begin{aligned}
        \widetilde{\mathcal{F}}_v &= \left( \mathcal{V}_{cls}^{k} | k = T - 1 \right) \\
        \mathcal{V}_{cls}^{k} = \mathrm{Focus} \, &\mathrm{Sampling}\left( \mathcal{F}_{v}^{k},\mathcal{V}_{cls}^{k-1} | \, bias \right) \\
    \end{aligned},
\end{equation}
where $k \in$ [$0,\,T-1$] and $T$ represents the total number of video 1s segments. The focus sampling is formulated as:
\begin{equation}
    \begin{aligned}
        &\mathcal{V}_{q}^{k-1} = \mathcal{F}_{v}^{k} + \mathcal{V}_{cls}^{k-1} \\
        &\mathcal{V}_{sa}^{k-1} = \mathrm{SAB}\left( \mathrm{SAB}\left( \mathcal{V}_{q}^{k-1} \right) + bias\right) + bias \\
        &\mathcal{V}_{cls}^{k} = \mathrm{CAB}\left(\mathcal{V}_{sa}^{k-1},\mathcal{F}_{v}^{k}\right) \\ 
    \end{aligned},
\end{equation}
where $\mathcal{V}_{sa}^{k-1}$ serves as $query$ and $\mathcal{F}_{v}^{k}$ serves as $key,value$ in $\mathrm{CAB}$. For the $k$-th time step audio feature $\mathcal{F}_{a}^{k}$, the same applies following the above Eqs.~(1-2).


\textit{(c)~Audio-visual key fusion:} After obtaining the fine-grained audio-visual focal features$\widetilde{\mathcal{F}}_a$ and $\widetilde{\mathcal{F}}_v$, we feed them into the audio-visual key fusion module, where multimodal fusion is performed under the guidance of the word-level question feature $\mathcal{F}_{t}^{w}$ to provide a refined semantic anchor for the subsequent decoding. The detailed audio-visual key fusion process is shown in Fig.~\ref{fig:3}~(c), $\widetilde{\mathcal{F}}_a$ and $\widetilde{\mathcal{F}}_v$ are transformed by linear layers and then added to the multimodal feature $\mathcal{F}_c$. Subsequently, each of these is fed into an SAB for enhancement, followed by a pooling operation, and then concatenated with $\mathcal{F}_c$ along the feature dimension. Finally, the result is passed through a linear layer and max sampling to obtain the final fused feature $\mathcal{F}_{fu}$. Where, $\mathcal{F}_c$ is formed by concatenating the visual feature $\mathcal{F}_v$, audio feature $\mathcal{F}_a$, and question feature $\mathcal{F}_t^{w}$ along the sequence length. The fusion process is formulated as follows:

\begin{equation}
    \begin{aligned}
        \mathcal{F}_c &= \mathrm{SAB}\left(\mathrm{Concat}\left(\mathcal{F}_a,\mathcal{F}_v,\mathcal{F}_t^w\right)\right) \\
        \mathcal{O}_a^l &= \mathrm{Pooling}\left(\mathrm{SAB}\left( \mathrm{Linear}\left(\widetilde{\mathcal{F}}_a \right) + \mathrm{Repeat}\left(\mathcal{F}_c \right) \right)\right) \\
        \mathcal{O}_v^l &= \mathrm{Pooling}\left(\mathrm{SAB}\left( \mathrm{Linear}\left(\widetilde{\mathcal{F}}_v \right) + \mathrm{Repeat}\left(\mathcal{F}_c \right)\right)\right) \\
        \mathcal{F}_{fu} &= \mathrm{Max}\left(\mathrm{Linear}\left(\mathrm{Concat}\left(\mathcal{O}_a,\mathcal{O}_v,\mathcal{F}_c\right)\right)\right) \\ 
    \end{aligned}
\end{equation}
where the $\mathrm{Pooling~(\cdot)}$ performs summation along the sequence dimension, while the $\mathrm{Max~(\cdot)}$ takes the maximum value along the feature dimension. The $\mathrm{Repeat~(\cdot)}$ is a broadcasting step along the sequence dimension and is used only for shape alignment before element wise addition. The fused feature $\mathcal{F}_{fu}$ $\in$ $\mathbb{R}^{1 \times D}$ and $\mathcal{O}$ represents the intermediate outputs.

\subsection{Global Preference Activation Path}\label{sec:3C}

Considering the potential mismatch between audio-visual segments and the possibility that the current question may be more biased toward a specific modality or scene in audio-visual question answering, we propose a global preference activation path. The preference activation path is designed to independently perceive and decouple auditory and visual inputs, activating the global contextual information within them. It serves to complement the fine-grained features from the temporal dynamic perception path, providing the model with an additional auxiliary perspective to achieve a more comprehensive understanding of the audio-visual scene.

As shown in Fig~\ref{fig:2}~(b), the visual feature $\mathcal{F}_v$ and the word-level question feature $\mathcal{F}_t^{w}$ are first fed into a cross-attention block. Meanwhile, the $\mathcal{F}_v$ is also separately enhanced via a self-attention block. The outputs are then summed and passed through a multi-layer perceptron~(MLP) to obtain the visual preference feature $\mathcal{F}_v^{p}$. Similarly, the audio feature $\mathcal{F}_a$ goes through the same procedure to obtain the audio preference feature $\mathcal{F}_a^{p}$. The calculation process is as follows:
\begin{equation}
    \begin{aligned}
        \mathcal{O}_v^g &= \mathrm{SAB}\left(\mathcal{F}_v\right) + \mathrm{CAB}\left(\mathcal{F}_v + \mathcal{F}_t^w \right) \\
        \mathcal{F}_v^{p} &= \mathrm{MLP}\left(\mathcal{O}_v^g\right) \\
        \mathcal{O}_a^g &= \mathrm{SAB}\left(\mathcal{F}_a\right) + \mathrm{CAB}\left(\mathcal{F}_a + \mathcal{F}_t^w \right) \\
        \mathcal{F}_a^{p} &= \mathrm{MLP}\left(\mathcal{O}_a^g\right) \\ 
    \end{aligned}
\end{equation}
where visual feature $\mathcal{F}_v$ or audio feature $\mathcal{F}_a$ serves as $query$ and $\mathcal{F}_t^w$ serves as $key,value$ in $\mathrm{CAB}$. The activated preference features $\mathcal{F}_v^{p}$ and $\mathcal{F}_a^{p}$ $\in$ $\mathbb{R}^{T \times D}$.


\begin{algorithm}[t!]
\caption{Temporal Dynamic Perception Path (TDPP)}
\label{alg:av-master-tdpp-extended}
\small
\textbf{Input:} \\
\hspace*{1em} Visual segment features $\mathcal{F}_v = \{\mathcal{F}_v^0, \dots, \mathcal{F}_v^{T-1}\} \in \mathbb{R}^{T \times D}$; \\
\hspace*{1em} Audio segment features $\mathcal{F}_a = \{\mathcal{F}_a^0, \dots, \mathcal{F}_a^{T-1}\} \in \mathbb{R}^{T \times D}$; \\
\hspace*{1em} Word-level question features $\mathcal{F}_t^w \in \mathbb{R}^{L \times D}$; \\
\hspace*{1em} Learnable temporal bias $b_{bias} \in \mathbb{R}^{T \times D}$. \\
\hspace*{1em} Predefined templates $\mathcal{V}_{cls}^{-1}, \mathcal{A}_{cls}^{-1} \in \mathbb{R}^{k \times D}$; \\
\textbf{Parameter:} \\
\hspace*{1em} The lengths of predefined audio-visual templates $k$ \\
\textbf{Output:} \\
\hspace*{1em} Fused collaborative audio-visual feature $\mathcal{F}_{fu}$.

\begin{algorithmic}[1]
\State \textbf{Stage 1: Audio-Visual Focus Capture}
\State \emph{Iteratively update templates to capture fine-grained cues:} 
\For{$k = 0$ \textbf{to} $T-1$} 
\Comment{Eq. 1, Fig. 3a}
    \State \emph{1.1 Visual Focus Sampling:}
    \State $\mathcal{V}_{q}^{k-1} \leftarrow \mathcal{V}_{cls}^{k-1} + \mathcal{F}_v^k$ \Comment{Inject current visual info}
    
    \State \emph{Self-Attention Block~(SAB) with Learned Bias:}
    \State $\mathcal{V}_{sa}^{k-1} \leftarrow \mathrm{MultiHeadAttn}(\mathbf{Q}=\mathcal{V}_{q}^{k-1}, \mathbf{K}=\mathcal{V}_{q}^{k-1}, \mathcal{V}_{q}^{k-1})$
    \State $\mathcal{V}_{sa}^{k-1} \leftarrow \mathrm{LayerNorm}(\mathcal{V}_{q}^{k-1} + \mathcal{V}_{sa}^{k-1} + b_{bias})$ \Comment{Eq. 2}
    
    \State \emph{Cross-Attention Block~(CAB) for Interaction:}
    \State $\mathcal{V}_{ca}^{k-1} \leftarrow \mathrm{MultiHeadAttn}(\mathbf{Q}=\mathcal{V}_{sa}^{k-1}, \mathbf{K}=\mathcal{F}_v^k, \mathbf{V}=\mathcal{F}_v^k)$
    \State $\mathcal{V}_{cls}^{k} \leftarrow \mathrm{LayerNorm}(\mathcal{V}_{sa}^{k-1} + \mathcal{V}_{ca}^{k-1})$ \Comment{Update template}
    \State \emph{1.2 Audio Focus Sampling:}
    \State $\mathcal{A}_{q}^{k-1} \leftarrow \mathcal{A}_{cls}^{k-1} + \mathcal{F}_a^k$ 
    \State $\mathcal{A}_{sa}^{k-1} \leftarrow \mathrm{MultiHeadAttn}(\mathbf{Q}, \mathbf{K},\mathbf{V}=\mathcal{A}_{q}^{k-1})$
    \State $\mathcal{A}_{sa}^{k-1} \leftarrow \mathrm{LayerNorm}(\mathcal{A}_{q}^{k-1} + \mathcal{A}_{sa}^{k-1} + b_{bias})$
    \State $\mathcal{A}_{ca}^{k-1} \leftarrow  \mathrm{MultiHeadAttn}(\mathcal{A}_{sa}^{k-1}, \mathcal{F}_a^k, \mathcal{F}_a^k)$
    \State $\mathcal{A}_{cls}^{k} \leftarrow \mathrm{LayerNorm}(\mathcal{A}_{sa}^{k-1} + \mathcal{A}_{ca}^{k-1})$ \Comment{Update template}
\EndFor
\State \emph{Final Fine-grained Focus Features:}
\State $\tilde{\mathcal{F}}_v \leftarrow \mathcal{V}_{cls}^{T-1}, \quad \tilde{\mathcal{F}}_a \leftarrow \mathcal{A}_{cls}^{T-1}$
\State \textbf{Stage 2: Audio-Visual Key Fusion}
\State \emph{Construct multimodal context anchor with question semantics:}
\State $\mathcal{C}_{multi} \leftarrow \mathrm{Concat}(\mathcal{F}_a, \mathcal{F}_v, \mathcal{F}_t^w)$
\State $\mathcal{F}_c \leftarrow \mathrm{SAB}(\mathcal{C}_{multi})$ \Comment{Eq. 3}

\State \emph{Fuse focus features with context:}
\State $\tilde{\mathcal{F}}_v^{proj} \leftarrow \mathrm{Linear}(\tilde{\mathcal{F}}_v), \quad \tilde{\mathcal{F}}_a^{proj} \leftarrow \mathrm{Linear}(\tilde{\mathcal{F}}_a)$
\State $\mathcal{O}_v \leftarrow \mathrm{Pooling}(\mathrm{SAB}(\tilde{\mathcal{F}}_v^{proj} + \mathcal{F}_c))$
\State $\mathcal{O}_a \leftarrow \mathrm{Pooling}(\mathrm{SAB}(\tilde{\mathcal{F}}_a^{proj} + \mathcal{F}_c))$

\State \emph{Final Aggregation via Max Sampling:}
\State $\mathcal{F}_{joint} \leftarrow \mathrm{Concat}(\mathcal{O}_a, \mathcal{O}_v, \mathcal{F}_c)$
\State $\mathcal{F}_{fu} \leftarrow \mathrm{Max}(\mathrm{Linear}(\mathcal{F}_{joint}))$ 

\State \Return $\mathcal{F}_{fu}$
\end{algorithmic}
\end{algorithm}

\subsection{Optimization and Answer Prediction}\label{sec:3D}

During training, the fused feature $\mathcal{F}_{fu}$ and the sentence-level question feature $\mathcal{F}_{t}^{s}$ are jointly fed into the multimodal decoder to generate the answer prediction distribution, which is then used together with the ground-truth labels to compute the answer loss (denoted as $\mathcal{L}_{qa}$). In the preference activation path, features $\mathcal{F}_{v}^{p}$ and $\mathcal{F}_{a}^{p}$ are separately input into two independent audio/visual decoders, which also generate prediction distributions under the guidance of $\mathcal{F}_{t}^{s}$, resulting in two preference losses~(denoted as $\mathcal{L}_{v}^{p}$ and $\mathcal{L}_{a}^{p}$). Moreover, a contrastive loss~(denoted as $\mathcal{L}_{c}$) is applied between the dynamic perception and preference activation paths, aiming to enhance the stability of the dual-path paradigm and improve the model's discriminative ability by leveraging hard negative samples. During the inference stage, we sum the prediction distributions generated by the multimodal decoder from the temporal dynamic perception path and the audio/visual decoders from the global preference activation path, and then apply the argmax function to obtain the final answer.

\textit{(a)~Dual-path prediction loss:} For the answer loss $\mathcal{L}_{qa}$, we follow the standard training procedure for the AVQA. The goal of the model is to minimize the negative log-likelihood of the probabilities over multiple answer choices generated by the multimodal decoder. For the preference losses $\mathcal{L}_{v}^{p}$ and $\mathcal{L}_{a}^{p}$, we also adopt a similar computation method to calculate them based on the probability distributions generated by the two audio/visual decoders. The formula for the above loss calculation can be represented as:
\begin{equation}
    \begin{aligned}
        \mathcal{L}_{qa} &= - \sum_{ans=1}^{C}y_{ans}\log\left(P_{ans}|\mathcal{F}_{fu},\theta_l\right) \\
        \mathcal{L}_{v}^{p}, \mathcal{L}_{a}^{p} &= - \sum_{pef=1}^{C}y_{pef}\log\left(P_{pef}|\mathcal{F}_{v}^{p},\mathcal{F}_{a}^{p},\theta_g\right) \\
    \end{aligned}
  \end{equation}
where $C$ is the total number of answer choices and $\theta$ is the set of learnable parameters of the decoders. Both the multimodal decoder and the audio/visual decoders include transformer blocks and linear layers.

\begin{table}[t]
    \centering
    \caption{Detailed description of AVQA, MUSIC-AVQA, MUSIC-AVQA-R, and MUSIC-AVQA-v2.0 datasets.} 
    \begin{adjustbox}{width=\linewidth,center=\linewidth}
    \begin{tabular}{l|c|c|c|c}
    \toprule
    \textbf{Dataset} & \textbf{\# Videos} & \textbf{\# Train QA} & \textbf{\# Valid QA} & \textbf{\# Test QA} \\ \hline
    AVQA        & 57,015           & 40,425    &  - & 16,910     \\
    MUSIC-AVQA        & 9,288           & 31,904    &  4,568 & 9,129     \\
    MUSIC-AVQA-R    & 9,288           &   -       &   -    & 211,572   \\ 
    MUSIC-AVQA-v2.0  & 10,492          & 37,408    &  5,346 & 10,819    \\ 
    \bottomrule 
    \end{tabular} 
    \end{adjustbox} 
    
    \label{tab:dataset_summary}
    \end{table}

\begin{table*}[t]
\centering
\caption{Experimental results on the MUSIC-AVQA test set. The best and second best performance of each task are highlighted in \textbf{bold} and \underline{underline} respectively. For a fair comparison, we report the performance of the version using the same audio encoder as our method, denoted as $\dagger$. Comparisons with other versions are presented in subsequent experiments.}
\begin{adjustbox}{width=\linewidth,center=\linewidth}
{
    \begin{tabular}{lccc|ccc|cccccc|c}
    \toprule 
        \multirow{2}{*}{\textbf{Methods}} & \multicolumn{3}{c|}{\textbf{Audio QA}} & \multicolumn{3}{c|}{\textbf{Visual QA}} & \multicolumn{6}{c|}{\textbf{Audio-Visual QA}} & \multirow{2}{*}{\textbf{Avg}} \\
        \cmidrule(lr){2-13}
         & \textbf{Count} & \textbf{Comp} & \textbf{Avg} & \textbf{Count} & \textbf{Local} & \textbf{Avg} & \textbf{Exist} & \textbf{Count} & \textbf{Local} & \textbf{Comp} & \textbf{Temp} & \textbf{Avg} & \\
        \cmidrule(lr){1-14}
        MCAN{\color{cyan!70}~\textit{[CVPR'19]}} & 77.50 & 55.24 & 69.25 & 71.56 & 70.93 & 71.24 & 80.40 & 64.91 & 54.48 & 57.22 & 47.57 & 61.58 & 65.49 \\
        PSAC{\color{cyan!70}~\textit{[AAAI'19]}} & 75.64 & 66.06 & 72.09 & 68.64 & 69.79 & 69.22 & 77.59 & 63.42 & 55.02 & 61.17 & 59.47 & 63.52 & 66.54 \\
        HME{\color{cyan!70}~\textit{[CVPR'19]}} & 74.76 & 63.56 & 70.61 & 67.97 & 69.46 & 68.76 & 80.30 & 63.19 & 53.18 & 62.69 & 59.83 & 64.05 & 66.45 \\
        AVSD{\color{cyan!70}~\textit{[CVPR'19]}} & 72.41 & 61.90 & 68.52 & 67.39 & 74.19 & 70.83 & 81.61 & 63.89 & 58.79 & 61.52 & 61.41 & 65.49 & 67.44 \\
        HCRN{\color{cyan!70}~\textit{[CVPR'20]}} & 68.59 & 50.92 & 62.05 & 64.39 & 61.81 & 63.08 & 54.47 & 53.38 & 41.53 & 52.11 & 47.69 & 50.26 & 55.73 \\
        Pano-AVQA{\color{cyan!70}~\textit{[ICCV'21]}} & 74.36 & 64.56 & 70.73 & 69.39 & 75.65 & 72.56 & 81.21 & 64.91 & 59.33 & 64.22 & 63.23 & 66.64 & 68.93 \\
        ST-AVQA{\color{cyan!70}~\textit{[CVPR'22]}} & 78.18 & 67.05 & 74.06 & 71.56 & 76.38 & 74.00 & 81.81 & 70.80 & 64.51 & \textbf{66.01} & 63.23 & 69.54 & 71.52 \\
        LAVISH{\color{cyan!70}~\textit{[CVPR'23]}} & 82.09 & 65.56 & 75.97 & 78.98 & 81.43 & 80.22 & 81.71 & 75.51 & 66.13 & 63.77 & 67.96 & 71.26 & 74.46 \\
        QAGL{\color{cyan!70}~\textit{[TCSVT'24]}} & 82.99 & \textbf{71.04} & \underline{78.58} & 80.12 & 77.88 & 78.89 & 82.29 & 72.73 & 62.83 & 63.40 & 64.36 & 69.43 & 73.58 \\
        TSPM{\color{cyan!70}~\textit{[ACMMM'24]}} & 84.07 & 64.65 & 76.91 & 82.29 & 84.90 & 83.61 & 82.19 & 76.21 & 71.85 & \underline{65.76} & \textbf{71.17} & 73.51 & 76.79 \\
        APL{\color{cyan!70}~\textit{[AAAI'24]}} & 82.40 & \underline{70.71} & 78.09 & 76.52 & 82.74 & 79.69 & 82.99 & 73.29 & 66.68 & 64.76 & 65.95 & 70.96 & 74.53 \\
        PSOT$^{\dagger }${\color{cyan!70}~\textit{[AAAI'25]}} & -- & -- & 78.22 & -- & -- & 80.07 & -- & -- & -- & -- & -- & 72.61 & 75.29 \\
        AVAF-Net{\color{cyan!70}~\textit{[AAAI'25]}} & 83.09 & 69.70 & 78.15 & 80.20 & 84.49 & 82.37 & \textbf{84.51} & 75.05 & 68.37 & 61.94 & 70.07 & 72.12 & 75.90 \\
        SHMamba{\color{cyan!70}~\textit{[TASLP'25]}} & 82.30 & 63.64 & 75.42 & 78.53 & 81.31 & 79.93 & 82.89 & 72.65 & 67.93 & 61.31 & 68.37 & 70.64 & 74.12 \\
        CoQo$^{\dagger }${\color{cyan!70}~\textit{[IJCV'25]}} & -- & -- & \underline{78.90} & -- & -- & 83.70 & -- & -- & -- & -- & -- & \underline{73.92} & 77.40 \\
        QA-TIGER{\color{cyan!70}~\textit{[CVPR'25]}} & \underline{84.86} & 67.85 & 78.58 & \underline{83.96} & \underline{86.29} & \underline{85.14} & 83.10 & \underline{78.58} & \textbf{72.50} & 63.94 & 69.59 & 73.74 & \underline{77.62} \\
        \rowcolor[HTML]{E6EAD9}
        \textbf{AV-Master~(Ours)} & \textbf{87.02} & 67.85 & \textbf{79.95} & \textbf{86.55} & \textbf{86.61} & \textbf{86.58} & \underline{83.60} & \textbf{79.13} & \underline{72.39} & 64.21 & \underline{70.80} & \textbf{74.22} & \textbf{78.51} \\
    \bottomrule
    \end{tabular}
}
\end{adjustbox}
\label{tab: music-avqa}
\end{table*} 

\textit{(b)~Dual-path contrastive loss:} For the contrastive loss $\mathcal{L}_{c}$, we enhance the stability of the dual-path architecture and improve the accuracy of joint prediction by increasing the similarity between the positive sample feature $\mathcal{F}_{fu}$ from the dynamic perception path and $\mathcal{F}_{g}^{j}$ from the preference activation path, $j \in \left\{v,a\right\}$. Meanwhile, we compare the feature $\mathcal{F}_{fu}$ from the positive sample with the feature $\mathcal{F}_{g}^{j}$ from the negative sample and reduce their similarity to enhance the model's ability to distinguish positive samples. The contrastive $\mathcal{L}_{c}$ is expressed as follows:
\begin{equation}
    \begin{aligned}
        &p_{1}^{j\left(i\right)} = exp\left(cos\left(\mathcal{F}_{fu}, \overline{\mathcal{F}}_{g}^{j\left(i\right)}\right)/\mathcal{T}\right) \\
        &p_{1}^{\left(i\right)} = p_{1}^{v\left(i\right)} + p_{1}^{a\left(i\right)} \\
        &\mathcal{L}_{c} = -\frac{1}{N} \sum_{i=1}^{N} \log \left[\frac{p_1^{\left(i\right)}}{\sum_{k \neq i}^{neg} p_1^{\left(k\right)} + p_1^{\left(i\right)}}\right] \\
    \end{aligned}
\end{equation}
where $neg$ is the number of negative pairs and $cos~(\cdot,\cdot)$ is the cosine function used to compute similarity. $\overline{\mathcal{F}}$ represents the operation of averaging the feature $\mathcal{F}$ along the sequence length dimension, $\mathcal{T}$ denotes the temperature. The overall loss is the weighted sum of the above losses:

\begin{equation}
    \begin{aligned}
        \mathcal{L} = \lambda_{qa} \mathcal{L}_{qa} + \lambda_{v}^{p} \mathcal{L}_{v}^{p} + \lambda_{a}^{p} \mathcal{L}_{a}^{p} + \lambda_{c} \mathcal{L}_{c} \\
    \end{aligned}
\end{equation}
where $\lambda_{qa}$, $\lambda_{v}^{p}$, $\lambda_{a}^{p}$, and $\lambda_{c}$ are hyperparameters used to trade-off each loss functions.

\section{Experiments}\label{sec:4}
\subsection{Experimental Setting}
\textit{(a)~Dataset and evaluation metric:} In this paper, we conduct experiments on four representative AVQA benchmarks: MUSIC-AVQA~\cite{li2022learning}, MUSIC-AVQA-R~\cite{ma2024look}, MUSIC-AVQA-v2.0~\cite{liu2024tackling}, and AVQA~\cite{yang2022avqa}, whose statistics are reported in Table~\ref{tab:dataset_summary}. MUSIC-AVQA is a widely used benchmark for audio-visual reasoning in music-performance scenes, containing 9,288 videos and 31,904/4,568/9,129 question-answer pairs for training, validation, and test, respectively. Since it is the most commonly adopted benchmark in prior work, we use it by default for our ablation and analysis experiments. MUSIC-AVQA-R is a robustness-oriented extension of MUSIC-AVQA, which keeps the same video set but expands the test split to 211,572 rephrased question-answer pairs, making it suitable for evaluating performance under more diverse question formulations and long-tail answer distributions. MUSIC-AVQA-v2.0 further improves the original benchmark by constructing a more balanced setting, and contains 10,492 videos with 37,408/5,346/10,819 question-answer pairs for training, validation, and test, respectively. We use it to assess the robustness of our method under reduced answer bias. To validate whether our model can generalize beyond music scenarios, we additionally evaluate on AVQA, a real-world benchmark containing 57,015 videos and 40,425/16,910 question-answer pairs for training/test. Following prior methods, we use answer accuracy (\%) as the main evaluation metric across all datasets.

\begin{table*}[t]
\renewcommand{\arraystretch}{1.05}
\renewcommand{\tabcolsep}{1.2mm}
\centering
\caption{Experimental results on the MUSIC-AVQA-R test set, with H and T representing performance on Head (frequent) and Tail (rare) answer categories, respectively. All results are obtained from official reports or reproduced from other works.}
\begin{adjustbox}{width=\linewidth,center=\linewidth}
\begin{tabular}{lcccc|cccc|cccccccccc|c}
\toprule
\multirow{4}{*}{\textbf{Methods}} & \multicolumn{4}{c|}{\textbf{Audio QA}} & \multicolumn{4}{c|}{\textbf{Visual QA}} & \multicolumn{10}{c|}{\textbf{Audio-Visual QA}} & \multirow{4}{*}{\textbf{Avg}} \\
\cmidrule(lr){2-19} 
& \multicolumn{2}{c}{\textbf{Count}} & \multicolumn{2}{c|}{\textbf{Comp}} & \multicolumn{2}{c}{\textbf{Count}} & \multicolumn{2}{c|}{\textbf{Local}} & \multicolumn{2}{c}{\textbf{Exist}} & \multicolumn{2}{c}{\textbf{Count}} & \multicolumn{2}{c}{\textbf{Local}} & \multicolumn{2}{c}{\textbf{Comp}} & \multicolumn{2}{c|}{\textbf{Temp}} & \\
\cmidrule(lr){2-19}
& \textbf{H} & \textbf{T} & \textbf{H} & \textbf{T} & \textbf{H} & \textbf{T} & \textbf{H} & \textbf{T} & \textbf{H} & \textbf{T} & \textbf{H} & \textbf{T} & \textbf{H} & \textbf{T} & \textbf{H} & \textbf{T} & \textbf{H} & \textbf{T} \\
\cmidrule(lr){1-20}
    HCAttn{\color{cyan!70}~\textit{[NeurIPS'16]}} &61.67 &41.63 &59.09 &47.14 &56.52 &9.20 &67.01 &53.16 &66.57 &61.13 &59.53 &12.48 &37.05 &42.48 &48.81 &60.12 &33.82 &39.26 &51.90 \\
    MCAN{\color{cyan!70}~\textit{[CVPR'19]}} &75.02 &60.16 &58.89 &50.09 &64.58 &26.69 &66.48 &62.25 &51.29 &67.29 &64.76 &25.28 &46.11 &61.61 &50.57 &52.40 &34.64 &58.05 &57.27 \\
    PSAC{\color{cyan!70}~\textit{[AAAI'19]}} &53.01 &56.68 &57.41 &48.12 &49.55 &26.43 &72.96 &60.69 &50.56 &55.54 &56.70 &19.58 &41.98 &52.30 &38.13 &58.92 &26.68 &46.24 &50.45 \\
    HME{\color{cyan!70}~\textit{[CVPR'19]}} &62.60 &53.95 &54.97 &\underline{58.29} &50.95 &16.46 &73.25 &58.60 &65.74 &66.49 &63.18 &17.18 &33.79 &46.03 &53.20 &69.57 &33.95 &41.57 &53.66 \\
    AVSD{\color{cyan!70}~\textit{[CVPR'19]}} &54.00 &47.84 &60.61 &47.79 &60.34 &10.07 &74.78 &61.43 &66.28 &61.98 &46.21 &8.06 &33.00 &40.35 &51.98 &66.00 &40.14 &41.52 &52.33  \\
    FCNLSTM{\color{cyan!70}~\textit{[TASLP'20]}} &66.23 &36.48 &64.78 &51.24 &61.75 &5.31 &54.86 &51.06 &64.76 &78.52 &62.69 &7.23 &46.66 &57.30 &43.13 &\underline{71.67} &37.02 &30.78 &54.12  \\
    HCRN{\color{cyan!70}~\textit{[CVPR'20]}} &55.53 &53.31 &47.17 &32.44 &41.87 &23.55 &39.40 &51.27 &41.81 &65.45 &54.58 &19.57 &36.62 &42.72 &33.33 &36.87 &40.47 &44.13 &43.92  \\
    Pano-AVQA{\color{cyan!70}~\textit{[ICCV'21]}} &50.57 &43.45 &50.78 &44.93 &47.28 &15.50 &67.19 &65.51 &52.37 &22.04 &52.21 &21.52 &44.35 &61.69 &45.61 &40.49 &35.00 &49.33 &47.40 \\
    ST-AVQA{\color{cyan!70}~\textit{[CVPR'22]}} &56.40 &41.48 &62.28 &57.59 &59.86 &12.94 &63.31 &54.00 &73.35 &77.26 &48.31 &8.41 &35.35 &40.49 &\textbf{53.30} &62.44 &40.25 &38.15 &52.80 \\
    LAVISH{\color{cyan!70}~\textit{[CVPR'23]}} &61.73 &43.99 &65.06 &\textbf{60.38} &65.53 &11.13 &70.21 &64.73 &\textbf{77.83} &\underline{79.46} &49.88 &14.87 &41.76 &41.20 &\underline{59.26} &65.10 &\underline{41.84} &46.26 &57.63 \\
    TSPM{\color{cyan!70}~\textit{[ACMMM'24]}} &81.65 &71.80 &67.66 &49.56 &78.29 &47.56 &80.58 &73.18 &69.15 &\textbf{82.79} &\underline{77.09} &\textbf{38.64} &42.24 &57.37 &52.07 &68.86 &39.23 &49.36 &66.30\\
    QA-TIGER{\color{cyan!70}~\textit{[CVPR'25]}} & \underline{82.67} & \textbf{75.82} & \textbf{71.75} & 43.11 & \underline{81.30} & \underline{54.59} & \underline{84.76} & \underline{75.59} & 72.84 & 78.56 & 76.70 & 33.55 & \underline{48.22} & \underline{64.65} & 37.55 & \textbf{80.47} & 36.85 & \underline{62.96} & \underline{67.99} \\
    \rowcolor[HTML]{E6EAD9}
    \textbf{AV-Master~(Ours)} & \textbf{84.90} & \underline{72.61} & \underline{70.67} & 49.22 & \textbf{83.48} & \textbf{57.40} & \textbf{87.39} & \textbf{79.33} & \underline{75.55} & 78.41 & \textbf{80.18} & \underline{35.28} & \textbf{55.39} & \textbf{77.41} & 47.76 & 70.80 & \textbf{46.49} & \textbf{69.99} & \textbf{71.19}\\
\bottomrule
\end{tabular}
\end{adjustbox}

\begin{flushleft}
\end{flushleft}
\label{tab: music-avqa-r}
\end{table*}

\begin{table}[t]
\renewcommand{\arraystretch}{1.05}
\renewcommand{\tabcolsep}{1.2mm}
\centering
\caption{Experimental results on the MUSIC-AVQA-v2.0 for (a) bias and (b) balanced test sets.}
\begin{adjustbox}{width=\linewidth,center=\linewidth}
\begin{tabular}{c|c|l|cccc}
\toprule
    \textbf{Test} & \textbf{Training} & \textbf{Methods} & \textbf{A-QA} & \textbf{V-QA} & \textbf{AV-QA} & \textbf{Avg} \\
    \hline 
    \multirow{10}{*}{\small (a) Bias} 
                                                & \multirow{4}{*}{Bias}        & ST-AVQA      & {76.86}     & 77.70          & 69.59          & 73.07          \\
                                                &                              & LAVISH       & 76.73          & {80.96}     & {70.80}     & {74.59}     \\
                                                &                              & QA-TIGER                & \underline{79.13} & \underline{84.83} & \underline{72.37} & \underline{76.93} \\
                                                &                              & \textbf{AV-Master}                & \textbf{79.31} & \textbf{86.54} & \textbf{74.12} & \textbf{78.39} \\ \cline{2-7}
                                                & \multirow{5}{*}{Balance}     & ST-AVQA      & 76.18          & 77.20          & 67.96          & 71.92          \\
                                                &                              & LAVISH       & 75.56          & 80.83          & 69.27          & 73.51          \\
                                                &                              & LAST        & {77.10}     & 82.99          & 70.86          & 75.24          \\
                                                &                              & LAST-Att    & \underline{77.29} & {83.47}     & {71.05}     & {75.45}     \\
                                                &                              & QA-TIGER                & 77.07          & \underline{85.93} & \underline{71.20} & \underline{76.57} \\
                                                &                              & \cellcolor[HTML]{E6EAD9} \textbf{AV-Master}                & \cellcolor[HTML]{E6EAD9} \textbf{79.25}          & \cellcolor[HTML]{E6EAD9} \textbf{86.87} & \cellcolor[HTML]{E6EAD9} \textbf{71.52} & \cellcolor[HTML]{E6EAD9} \textbf{77.03} \\
    \toprule
    \addlinespace[0.3em]
    \bottomrule
    \textbf{Test} & \textbf{Training} & \textbf{Methods} & \textbf{A-QA} & \textbf{V-QA} & \textbf{AV-QA} & \textbf{Avg} \\
    \hline
    \multirow{10}{*}{\small (b) Balance} 
                                                & \multirow{4}{*}{Bias}        & ST-AVQA   & {73.34}     & 76.82          & 64.51          & 69.40          \\
                                                &                              & LAVISH    & 73.14          & {79.70}     & {65.01}     & {70.39}     \\
                                                &                              & QA-TIGER             & \underline{77.57} & \underline{84.84} & \underline{67.43} & \underline{73.91} \\ 
                                                &                              & \textbf{AV-Master}             & \textbf{78.22} & \textbf{86.42} & \textbf{69.11} & \textbf{75.37} \\ \cline{2-7}
                                                & \multirow{6}{*}{Balance}     & ST-AVQA   & 75.50          & 77.67          & 66.32          & 71.02          \\
                                                &                              & LAVISH    & 76.15          & 81.32          & 68.28          & 73.18          \\
                                                &                              & LAST     & 78.08          & 83.29          & 69.72          & 74.85          \\
                                                &                              & LAST-Att & {78.56}     & {84.07}     & \textbf{70.30} & {75.44}     \\
                                                &                              & QA-TIGER             & \underline{79.90} & \underline{86.95} & {70.22}     & \underline{76.43} \\ 
                                                &                              & \cellcolor[HTML]{E6EAD9} \textbf{AV-Master}            & \cellcolor[HTML]{E6EAD9} \textbf{80.84} & \cellcolor[HTML]{E6EAD9} \textbf{87.37} & \cellcolor[HTML]{E6EAD9} \underline{70.29}     & \cellcolor[HTML]{E6EAD9} \textbf{76.75} \\
\bottomrule
    \end{tabular}
\end{adjustbox}

\label{tab: music-avqa-v2}
\end{table}

\begin{table}[t]
    \centering
    \caption{Experimental results on the test set of AVQA dataset.} 
    \begin{adjustbox}{width=\linewidth,center=\linewidth}
    \begin{tabular}{c|c|c}
    \toprule
    \multirow{1}{*}{\textbf{Methods}} & \multirow{1}{*}{\textbf{Ensemble}} & \multirow{1}{*}{\textbf{Total Accuracy (\%)}} \\ \hline 
    HME    &  HAVF & 85.0      \\ 
    PSAC   &  HAVF & 87.4    \\ 
    LADNet &  HAVF & 84.1  \\ 
    ACRTransformer &  HAVF & 87.8       \\ 
    HGA &  HAVF & 87.7       \\ 
    HCRN &  HAVF & 89.0       \\ \hline
    SaSR-Net &  -- & 89.9       \\
    PSTP-Net &  -- & 90.2       \\
    TSPM &  -- & 90.8       \\
    MCD &  -- & 90.8       \\ \rowcolor[HTML]{E6EAD9}
    \textbf{AV-Master~(Ours)} & -- & \textbf{91.4} \\ \bottomrule
    \end{tabular}
    \end{adjustbox}
    \label{tab:avqa}
    \end{table}

\begin{table}[t]
\renewcommand{\arraystretch}{1.5}
\centering
\caption{Comparison with pretraining-based methods. Z-S indicates whether the zero-shot setting is used and PT represents the amount of data used for model pre-training.}
\begin{adjustbox}{width=\linewidth,center=\linewidth}
{\begin{tabular}{lcccccc}
\toprule
{\textbf{Methods}} &   {\textbf{V-Enc.}} & {\textbf{A-Enc.}} & {\textbf{Z-S}}                   & {\textbf{PT}} & {\textbf{Params}} & {\textbf{ACC}}  \\
\toprule
{OneLLM} & {CLIP$_{\text{L}}$}  &   {CLIP$_{\text{L}}$} & {\checkmark}                             & 1008.5M                  & 7B                    & 47.6                                \\
{ChatBridge} & {ViT$_{\text{G}}$}  &   {BEAT} & {\checkmark}                             & 130.0M                  & 13B                    & 43.0                                \\
{CAT} & {lmageBind}  &   {Imagebind} & {\checkmark}                             &        3.1M           & 7B                    & 48.6                                \\
{CAT+} & {lmageBind}  &   {Imagebind} & {\checkmark}                             &         0.2M          & 7B                    & 50.1                                \\
{VideoLLaMa} & {EVACLIP$_{\text{G}}$}  &   {Imagebind} & {\checkmark}                             & 2.8M                  & 7B                    & 36.6                                \\
{AVLLM} & {CLIP$_{\text{L}}$}    &   {CLAP} & {\checkmark}                        & 1.6M                 & 13B                    & 45.2                                   \\
{AVicuna} & {CLIP$_{\text{L}}$}        &   {CLAP} & {\checkmark}                                      & 1.1M                 & 7B                    & 49.6                                  \\
\hline \hline
{CAD} & {{ViT}} &   {PANNs} & {\ding{53}}                                     & 100.0M                  & - -                    & 78.3                                 \\
{VAST} & {{EVACLIP$_{\text{G}}$}} &   {BEATs} & {\ding{53}}                                     & 42.0M                  & 1.3B                    & 80.7                                 \\
{VALOR} & {{CLIP$_{\text{L}}$}} &   {AST} & {\ding{53}}                                     & 33.5M                  & 593M                    & 78.9                                 \\
\rowcolor[HTML]{E6EAD9}
{\textbf{AV-Master}} & {{CLIP$_{\text{L}}$}}  &   {VGGish} & {\ding{53}}                             & N/A                  & 34M                    & 78.5                                 \\
                         \bottomrule
\end{tabular}}
\end{adjustbox}
\label{tab:avllm}
\end{table}

\begin{table}[t]
    \centering
    \caption{ Parameters and computational complexity comparison with representative methods.}
    \label{tab:Parameters}
    \begin{adjustbox}{width=\linewidth,center=\linewidth}
    {\begin{tabular}{c|ccc|c}
    \toprule
     \multirow{1}{*}{\textbf{Methods}} & \multicolumn{1}{c}{ \textbf{Params}} & \multirow{1}{*}{\textbf{FLOPs}} & \multirow{1}{*}{\textbf{Latency}} & \multirow{1}{*}{\textbf{Acc~(\%)}} \\   
    \hline
     {ST-AVQA}           & 18.48M                            & 3.19G                   & 47.62ms                    & 71.59                                \\
     {QA-Tiger}         & 63.09M                                     & 9.57G                  & 63.28ms                    & 77.62                  \\
     \hline
     {AV-Master}          & 34.07M                                     & 6.20G                  & 89.24ms                    & 78.51                  \\
                                     \bottomrule
    \end{tabular}}
    \end{adjustbox}
\end{table}

\textit{(b)~Implementation details:} For a fair comparison, we follow previous work~\cite{kim2025question} and adopt a similar setup: videos are uniformly sampled at a rate of 1 frame per second. Audio representations are extracted using the pre-trained VGGish model~\cite{hershey2017cnn}, while visual inputs and corresponding questions are encoded through the CLIP-ViT-L/14 model~\cite{radford2021learning}. All extracted features are projected into a 768-dimensional space via a linear transformation.  The model is optimized using Adam~\cite{kingma2014adam} and starts with a learning rate of 1e-4, which is reduced by a factor of 0.1 every 8 epochs. The batch size is set to 64, and the model is trained for 30 epochs. Our proposed model is trained on NVIDIA GeForce RTX 4090 and implemented in PyTorch.


\textit{(c)~Predefined template initialization:} The predefined audio and visual templates are initialized in a data-dependent manner rather than from random noise. After extracting the audio and visual segment features, all features are projected into a 768-dimensional space. We set the length of both the audio and visual templates to 8. Specifically, the initial visual template ($ \mathcal{V}_{cls}^{-1}$) is constructed from the visual feature of the first time step ($ \mathcal{F}_v^0$), while the initial audio template ($\mathcal{A}_{cls}^{-1}$) is constructed from the audio feature of the first time step ($\mathcal{F}_a^0$). The first-step features are expanded along the template dimension to match the predefined template length. These templates serve as modality-specific initial anchors for the subsequent dynamic adaptive focus sampling process. During temporal updating, the current segment feature is injected into the previous template, and the template is progressively refined through self-attention, learnable temporal bias, and cross-attention. After processing all time steps, the final templates ($\mathcal{V}_{cls}^{T-1}$) and ($\mathcal{A}_{cls}^{T-1}$) are taken as the fine-grained visual and audio focus features, respectively.

\subsection{Quantitative Results}

\begin{table}[t]
    \centering
    \caption{Ablation on the different components of AV-Master.}
    \begin{adjustbox}{width=\linewidth,center=\linewidth}
    \begin{tabular}{c|c|cccc}
    \toprule
    \multirow{2}{*}{\textbf{\#}} & \multirow{2}{*}{\textbf{Methods}} & \multicolumn{4}{c}{\textbf{Average Accuracy (\%)}} \\ \cline{3-6}
    & & \textbf{A-QA} & \textbf{V-QA} & \textbf{AV-QA} & \textbf{Avg} \\ \hline
    1 &  w/o. ALL & 73.37 & 79.23 & 69.82 & 72.94 \\ 
    2 &  w/o. AVFC & 79.83 & 85.55 & 74.04 & 78.11 \\ 
    3 &  w/o. AVKF & 79.76 & 85.84 & 74.06 & 78.19 \\
    4 &  w/o. DPCL & 78.83 & 85.71 & 73.78 & 77.84 \\ 
    5 &  w/o. GPAP & 77.90 & 85.05 & 73.10 & 77.12 \\ 
    6 &  w/o. TDPP & 78.77 & 83.73 & 72.94 & 76.83 \\ \hline
    7 &  Dynamic Fus & 77.53 & 84.97 & 73.78 & 77.41 \\ 
    8 & \textbf{AV-Master} & \textbf{79.95} & \textbf{86.58} & \textbf{74.22} & \textbf{78.51} \\ \bottomrule
    \end{tabular}
    \end{adjustbox}
    \label{tab:ab}
    \end{table}


\begin{table}[t]
    \centering
    \caption{Ablation study of different loss functions in AV-Master training.}
    \label{tab:loss}
    \begin{adjustbox}{width=\linewidth,center=\linewidth}
    {\begin{tabular}{c|cccc|cccc}
    \toprule
    \multirow{2}{*}{\textbf{\#}} & \multirow{2}{*}{\textbf{$\mathcal{L}_{qa}$}} & \multirow{2}{*}{\textbf{$\mathcal{L}_{c}$}} & \multirow{2}{*}{\textbf{$\mathcal{L}_{a}^{p}$}} & \multirow{2}{*}{\textbf{$\mathcal{L}_{v}^{p}$}} & \multicolumn{4}{c}{\textbf{Average Accuracy (\%)}} \\ \cline{6-9}
    & & & & & \textbf{A-QA} & \textbf{V-QA} & \textbf{AV-QA} & \textbf{Avg} \\
    \hline
    {1} & {\checkmark} & {} & {} & {}     & 77.84                        & 84.31                 & 71.11                    & 75.80                                \\
    {2} & {\checkmark} & {\checkmark} & {} & {}     & 78.15                                      & 84.43                 & 70.76                   & 75.69    \\
    {3} & {\checkmark} & {\checkmark}   & {\checkmark} & {}  & 79.83                                     & 82.74                  & 72.49                    & 76.50    \\
    {4} & {\checkmark} & {\checkmark}  & {} & {\checkmark}   & 78.83                                     & 86.33                  & 72.63                    & 77.36    \\
    {5} & {\checkmark} & {}  & {\checkmark} & {\checkmark}   & 78.83                                     & 85.71                  & 73.78                    & 77.84    \\ 
    {6} & {\checkmark} & {\checkmark}   & {\checkmark} & {\checkmark}  & \textbf{79.95} & \textbf{86.58} & \textbf{74.22} & \textbf{78.51} \\   

                                     \bottomrule
    \end{tabular}}
    \end{adjustbox}
  \end{table}

\textit{(a)~Compared methods:} To evaluate the effectiveness and superiority of our model, we compare AV-Master with existing state-of-the-art AVQA methods across multiple datasets. These methods include: QA-TIGER~\cite{kim2025question}, CoQo~\cite{pei2025guiding}, SHMamba~\cite{yang2025shmamba}, AVAF-Net~\cite{zhao2025audio}, PSOT~\cite{li2025patch},  SaSR-Net~\cite{yang2024sasr}, PSTP-Net~\cite{li2023progressive}, TSPM~\cite{li2024boosting}, APL~\cite{li2024object}, LAST~\cite{liu2024tackling}, MCD~\cite{ye2024answering}, QAGL~\cite{chen2023question}, LAVISH~\cite{lin2023vision}, ST-AVQA~\cite{li2022learning}, Pano-AVQA~\cite{yun2021pano}, ACRTransformer~\cite{zhang2020action}, HGA~\cite{jiang2020reasoning}, HCRN~\cite{le2020hierarchical}, FCNLSTM~\cite{fayek2020temporal}, LADNet~\cite{li2019learnable}, AVSD~\cite{schwartz2019simple}, HME~\cite{fan2019heterogeneous}, PSAC~\cite{li2019beyond}, MCAN~\cite{yu2019deep} and HCAttn~\cite{lu2016hierarchical}.

\textit{(b)~MUSIC-AVQA:} As shown in Table~\ref{tab: music-avqa}, AV-Master achieves an overall accuracy of 78.51\%, outperforming all existing models, including the state-of-the-art method QA-TIGER~(77.62\%). Notably, our model demonstrates strong performance on complex reasoning tasks such as counting, significantly outperforming the second-best method~(i.e., A-Counting: 87.02\% vs. 84.86\%, V-Counting: 86.55\% vs. 83.96\%, AV-Counting: 79.13\% vs. 78.58\%).

\textit{(c)~MUSIC-AVQA-R:} As shown in Table~\ref{tab: music-avqa-r}, compared to the MUSIC-AVQA dataset, AV-Master demonstrates more significant improvements~(+3.20\%) on the MUSIC-AVQA-R dataset, achieving an overall accuracy of 71.19\% and setting a new state-of-the-art performance. This significant performance gain can be attributed to the dual-path learning paradigm of AV-Master. Since MUSIC-AVQA-R substantially increases question diversity through large-scale rephrasing and further evaluates performance under answer head–tail imbalance, it favors models with stronger semantic grounding rather than those relying on fixed question templates. AV-Master benefits from this setting, as its TDPP builds answers upon temporally continuous evidence that is relevant to the question, while GPAP provides stable modality-level priors directly from raw unimodal features. As a result, our model is more robust to rephrased questions and distribution shifts, leading to greater improvements on MUSIC-AVQA-R.


\textit{(d)~MUSIC-AVQA-v2.0:} As shown in Table~\ref{tab: music-avqa-v2}, AV-Master outperforms existing models across all types. Notably, when trained on the biased dataset, AV-Master still achieves significant improvements over the second-best method on both the balanced and biased test sets. These results highlight the robustness and adaptability of AV-Master in handling various training environments, demonstrating its strong generalization capability even under distributional bias.


\begin{table}[t]
    \centering
    \caption{Impact of lengths of audio-visual templates. A-L and V-L denote the lengths of the audio template and the visual template, respectively.}
    \begin{adjustbox}{width=\linewidth,center=\linewidth}
    \begin{tabular}{c|c|c|cccc}
    \toprule
    \multirow{2}{*}{\textbf{\#}} & \multirow{2}{*}{\textbf{A-L}} & \multirow{2}{*}{\textbf{V-L}} & \multicolumn{4}{c}{\textbf{Average Accuracy (\%)}} \\ \cline{4-7}
    & & & \textbf{A-QA} & \textbf{V-QA} & \textbf{AV-QA} & \textbf{Avg} \\ \hline
    1 & 16 & 4 & 80.01 & 85.67 & 73.76 & 78.03  \\ 
    2 & 12 & 8 & 79.70 & 86.17 & 73.88 & 78.17\\ 
    3 & 8 & 12  & 80.32 & 85.96 & 73.94 & 78.26  \\ 
    4 & 4 & 16  & 80.44 & 85.30 & 73.98 & 78.12 \\
    \hline
    5 & 16 & 16 & 80.51 & 85.96 & 74.04 & 78.34  \\ 
    6 & 12 & 12 & 80.26 & 85.88 & 74.08 & 78.30 \\ 
    7 & 8 & 8  & 79.95 & \textbf{86.58} & \textbf{74.22} & \textbf{78.51} \\ 
    8 & 4 & 4  & \textbf{80.63} & 85.22 & 73.92 & 78.10 \\ 
    9 & 2 & 2  & 80.38 & 85.84 & 73.98 & 78.26 \\ 
    \bottomrule
    \end{tabular}
    \end{adjustbox}
    \label{tab:lengths}
    \end{table}

\begin{table}[t]
    \centering
    \caption{Impact of weight sharing strategies. A-S and B-S represent attention block weight sharing and learned bias weight sharing, respectively.}
    \label{tab:weight_sharing}
    \begin{adjustbox}{width=\linewidth,center=\linewidth}
    {\begin{tabular}{c|c|c|cccc}
    \toprule
    \multirow{2}{*}{\textbf{\#}} & \multirow{2}{*}{\textbf{A-S}} & \multirow{2}{*}{\textbf{B-S}} & \multicolumn{4}{c}{\textbf{Average Accuracy (\%)}} \\ \cline{4-7}
    & & & \textbf{A-QA} & \textbf{V-QA} & \textbf{AV-QA} & \textbf{Avg} \\
    \hline
    {1} & {} & {}      & 79.70                        & 85.80                 & 73.92                    & 78.09                                \\
    {2} & {\checkmark} & {}      & 79.95                                      & \textbf{86.58}                 & \textbf{74.22}                    & \textbf{78.51}    \\
    {3} & {} & {\checkmark}     & \textbf{80.26}                                     & 85.88                  & 73.59                    & 78.03    \\
    {4} & {\checkmark} & {\checkmark}     & 80.07                                     & 86.29                  & 73.88      & 78.27    \\

                                     \bottomrule
    \end{tabular}}
    \end{adjustbox}
  \end{table}


\textit{(e)~AVQA:} To further validate the generalization capability of our model in real-world scenarios, we conducted experiments on the AVQA dataset. As shown in Tab.~\ref{tab:avqa}, AV-Master achieved an overall accuracy of 91.4\%, surpassing all previous methods both with and without the HAVF~\cite{yang2022avqa} module. These results strongly demonstrate that AV-Master maintains its exceptional performance in complex, real-world audio-visual question answering tasks. It is worth noting that although the performance improvement of AV-Master on the AVQA dataset may seem limited compared to its performance on the MUSIC-AVQA series of datasets, this is primarily due to the shorter duration and simpler audio-visual content of the AVQA dataset.

\begin{figure}[t]
    \centering
    \includegraphics[width=0.65\linewidth, keepaspectratio]{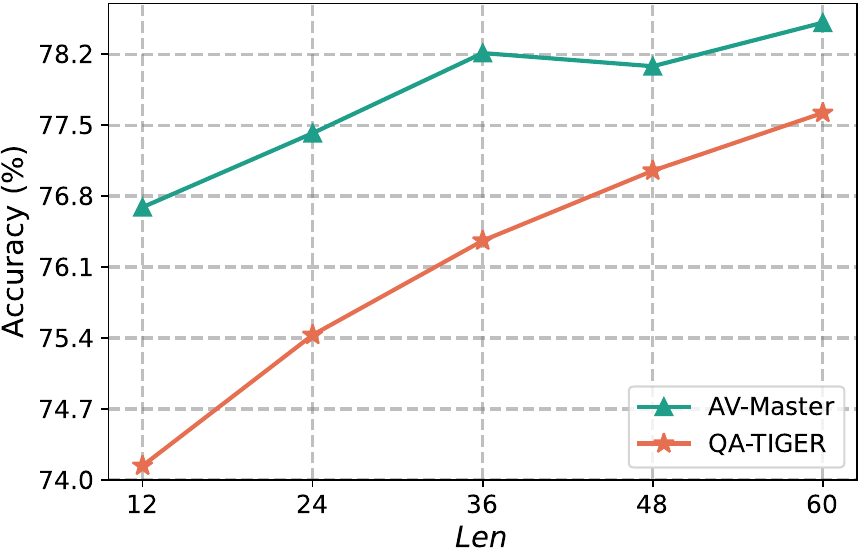}
    \vspace{-0.7em}
    \caption{The ablation study on the lengths of visual and audio segments and comparison with QA-TIGER.}
    \label{fig:9}
\end{figure}

\begin{table}[t]
  \centering
  \caption{Different visual and audio feature extractors.}
  \begin{adjustbox}{width=0.9\linewidth,center=\linewidth}
  {\begin{tabular}{c|c|c|cccc}
  \toprule
\multirow{2}{*}{\makecell[c]{\textbf{Visual} \\ \textbf{Encoder}}} & \multirow{2}{*}{\makecell[c]{\textbf{Audio} \\ \textbf{Encoder}}} & \multirow{2}{*}{\textbf{Methods}} & \multicolumn{4}{c}{\textbf{Average Accuracy (\%)}} \\ \cline{4-7}
  & & & \textbf{A-QA} & \textbf{V-QA} & \textbf{AV-QA} & \textbf{Avg} \\
  \hline
  \multirow{2}{*}{Resnet-18} & \multirow{2}{*}{VGGish}  & {ST-AVQA}   & 74.06                                      & 74.00                 & 69.54                    & 71.52        \\
  & & {\textbf{AV-Master}}           & \textbf{78.21}                             & \textbf{79.23}                  & \textbf{70.25}                    & \textbf{74.04}                                \\
  \hline
  \multirow{2}{*}{CLIP$_{\text{B}}$} & \multirow{2}{*}{VGGish} & {PSTP-Net}           & 70.91                             & 77.26                  & 72.57                    & 73.52                                \\
  & & {\textbf{AV-Master}}           & \textbf{78.40}                             & \textbf{82.25}                  & \textbf{72.82}                    & \textbf{76.31}                                \\
  \hline
  \multirow{5}{*}{CLIP$_{\text{L}}$} & \multirow{3}{*}{VGGish} & {TSPM}         & 76.91                                     & 83.61                  & 73.51                    & 76.79                                \\
  & & {QA-TIGER}         & 78.58                                     & 85.14                  & 73.74                    & 77.62                                \\
  & & \textbf{{AV-Master}}          & \textbf{79.95}                                     & \textbf{86.58}                  & \textbf{74.22}                    & \textbf{78.51}                                 \\
  \cline{2-7}
   & \multirow{2}{*}{CLAP} & {PSOT}         & 79.08                                     & 87.12                  & 74.07                    & 78.42                                \\
  & & \textbf{{AV-Master}}          & \textbf{80.63}                                     & \textbf{87.78}                  & \textbf{74.92}                    & \textbf{79.34}                                 \\
  \hline
  \multirow{2}{*}{Internvideo2} & \multirow{2}{*}{Internvideo2} & {CoQo}         & 79.27                                     & 87.90                  & 75.80                    & 79.60                                \\
  & & \textbf{{AV-Master}}          & \textbf{81.50}                                     & \textbf{88.32}                  & \textbf{75.86}                    & \textbf{80.15}                                 \\

                           \bottomrule
  \end{tabular}}
\end{adjustbox}
  \label{tab:encoder}
\end{table}

\begin{figure}[t]
    \centering
    \includegraphics[width=\linewidth, keepaspectratio]{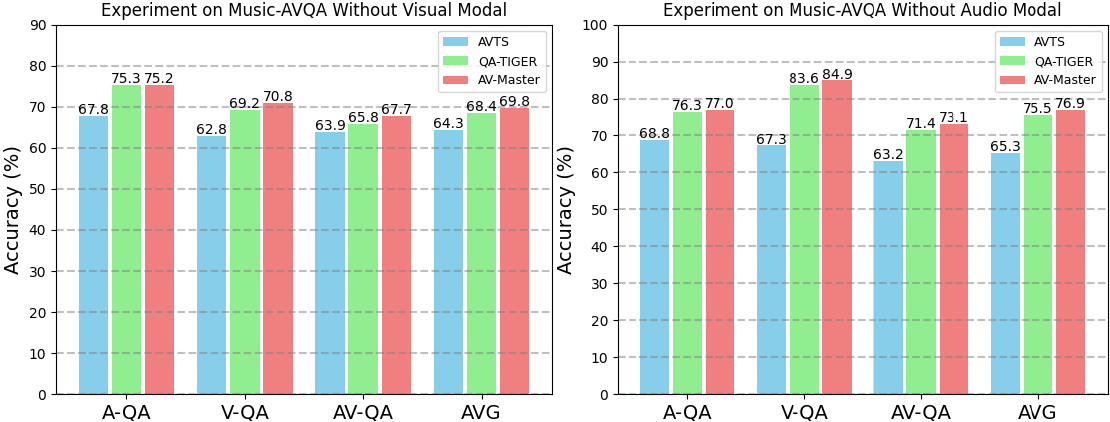}
    \caption{The ablation study on input modalities and comparison with other popular models~(AVTS and QA-TIGER).}
    \label{fig:6}
\end{figure}

\textit{(f)~Comparison with pretraining-based methods:} In this work, our approach focuses on designing efficient AVQA expert models that achieve competitive performance under limited training data and hardware conditions by developing powerful modules. This is also the mainstream direction in current AVQA research~\cite{kim2025question, pei2025guiding, yang2024sasr}. Additionally, there are some methods~\cite{damonlpsg2024videollama2, chen2023vast, chen2023valor} that rely on large-scale pretraining, which attempt to apply large language models to audio-visual scenarios to handle downstream AVQA tasks. However, pretraining-based AVQA methods require substantial computational resources and multimodal data, and their generalization ability in specific audio-visual scenarios falls short of expectations. We present relevant experiments to compare AV-Master with pretraining-based models, including OneLLM~\cite{han2024onellm}, Chatbridge~\cite{zhao2023chatbridge}, CAT~\cite{ye2024cat}, CAT+~\cite{ye2025cat+}, VideoLLaMa~\cite{zhang2023video}, AVLLM~\cite{shu2023audio}, AVicuna~\cite{tang2024avicuna}, VideoLLaMa2~\cite{damonlpsg2024videollama2}, CAD~\cite{nadeem2024cad}, VAST~\cite{chen2023vast} and VALOR~\cite{chen2023valor}. 

As shown in Table~\ref{tab:avllm}, AV-Master achieves competitive performance~(78.5\% accuracy) compared to various pretraining-based models, despite having significantly fewer parameters~(34M) and not relying on large-scale pretraining. In contrast, methods such as CAT+ and AVicuna, which are based on large language models, achieve notably lower performance~(50.1\% and 49.6\% accuracy) under zero-shot conditions. Although these approaches leverage abundant resources and powerful backbone models, they often exhibit weaker generalization capabilities in domain-specific audio-visual tasks. Other pretraining-based models achieve impressive performance (80.7\% accuracy) after fine-tuning on the MUSIC-AVQA dataset. However, our method achieves comparable results using significantly less training data, fewer trainable parameters, and a more lightweight audio-visual feature encoder. This highlights the practicality of AV-Master and its suitability for resource-constrained audio-visual scenarios.

It also demonstrates that model size and the scale of pretraining data are not the only factors influencing AVQA performance. Although large pretrained models can learn broad cross-modal alignments, AVQA additionally requires fine-grained temporal localization and question-relevant modality selection. AV-Master introduces these task-specific inductive biases through the temporal dynamic perception path and the global preference activation path. As a result, even with only 34M parameters and without additional pretraining, AV-Master remains competitive with larger pretrained methods, highlighting the importance of architecture design for AVQA.


\textit{(g)~Comparison of parameters and computational complexity:} To further evaluate whether the performance improvement of AV-Master comes at the cost of efficiency, we compare it with representative methods in terms of parameter count, FLOPs, inference latency, and accuracy, as shown in Table~\ref{tab:Parameters}. AV-Master achieves the best accuracy of 78.51\% among the compared methods. Compared to ST-AVQA, AV-Master achieves a substantial performance gain (+6.92\%) at the cost of acceptable overhead. We believe that this significant improvement more than justifies the marginal loss in efficiency. Compared with QA-TIGER, our method uses substantially fewer parameters (34.07M vs. 63.09M) and lower FLOPs (6.20G vs. 9.57G), while still achieving higher accuracy (78.51\% vs. 77.62\%). This suggests that the performance gain of AV-Master does not stem from simply scaling up the model size, but from the effectiveness of the proposed dual-path design.

Although AV-Master has fewer parameters and lower FLOPs than QA-TIGER, its latency is higher because FLOPs do not fully reflect practical GPU execution time. In AV-Master, the temporal dynamic perception path updates audio and visual templates sequentially over time, where the template at time step k depends on that from time step k-1. This recurrent focus-sampling process limits temporal parallelization and introduces extra attention, normalization, and memory-access overhead. In addition, AV-Master uses a dual-path inference strategy with multimodal, audio, and visual decoders, which further increases runtime. Therefore, the higher latency mainly comes from sequential temporal updating and multi-branch decoding rather than model size.

\subsection{Ablation Studies}

\textit{(a)~Ablation study on main components:} To explore the effectiveness of each component, we removed them individually and re-evaluated the performance. As shown in Tab.~\ref{tab:ab}, removing different components leads to varying degrees of performance degradation for AV-Master. Specifically, removing the TDPP (Temporal Dynamic Perception Path) causes the largest drop in overall accuracy, from 78.51\% to 76.83\%, indicating that fine-grained temporal perception is the most critical source of performance gain in our framework. Removing the GPAP (Global Preference Activation Path) also results in a clear decrease to 77.12\%, which verifies the importance of modeling modality preference from a global perspective. In addition, removing the DPCL (Dual-path Contrastive Loss) lowers the performance to 77.84\%, showing that collaborative optimization between the two paths is beneficial for learning more discriminative cross-modal representations.

We further analyze the contributions of the two modules inside the temporal dynamic perception path. Removing AVFC (Audio-Visual Focus Capture) reduces the average accuracy to 78.11\%, while removing AVKF (Audio-Visual Key Fusion) yields 78.19\%. These results suggest that AVFC and AVKF are complementary: the former is responsible for extracting fine-grained question-relevant cues from the original audio-visual sequence, whereas the latter further integrates the focused audio-visual features with multimodal context to produce a more discriminative fused representation. When all designed components are removed simultaneously, the performance drops substantially from 78.51\% to 72.94\%, confirming that the overall improvement comes from the joint contribution of all proposed modules rather than any single design alone.

Moreover, to further compare our modality modeling strategy with a representative late-fusion approach, we introduce a Dynamic Fus variant implemented based on a question-oriented audio-visual adaptive fusion strategy. This variant achieves an average accuracy of 77.41\%, which is 1.10\% lower than AV-Master. The performance gap is particularly evident on the A-QA and V-QA sub-tasks, where AV-Master improves from 77.53\% to 79.95\% and from 84.97\% to 86.58\%, respectively. These results indicate that assigning dynamic modality weights only at the fusion stage is less effective than our preference enhancement strategy. Our approach directly models modality preference from the initial unimodal features. By activating question-relevant global information before multimodal entanglement, GPAP provides more reliable modality guidance, which complements the fine-grained temporal cues captured by TDPP.

\textit{(b)~Impact of template lengths:} We further investigate the impact of predefined audio and visual template lengths on model performance by independently varying the lengths of $\mathcal{A}_{cls}$ and $\mathcal{V}_{cls}$. As shown in Table~\ref{tab:lengths}, symmetric and moderate-length template configurations are more favorable. For example, the model achieves the highest average accuracy when both template lengths are set to 8. A possible reason is that our AVFC module adopts a parameter-sharing strategy across modalities. Using balanced template lengths helps maintain similar representational capacity in the audio and visual branches, making the shared focus-sampling process more stable and effective. In contrast, overly short templates may miss useful cues, while overly long or highly imbalanced templates may introduce redundant information and weaken cross-modal consistency, leading to degraded performance.

\textit{(c)~Impact of weight sharing strategies:} We also explored the impact of different weight-sharing strategies in focus sampling. As shown in Table~\ref{tab:weight_sharing}, the model achieves the best overall performance when attention blocks share weights while biases remain unshared — this configuration is also adopted as our default setting. Interestingly, when only biases are shared, the model attains the lowest overall accuracy but achieves the highest accuracy on the A-QA task. This suggests that bias sharing may benefit certain subtasks while potentially hindering overall model performance.

\textit{(d)~Ablation study on loss functions:} To analyze the contribution of each loss function, we conducted an ablation study with the results presented in the Tab.~\ref{tab:loss}. This study systematically evaluates the model's performance by incrementally adding different loss components. The baseline model, trained only with the answer loss $\mathcal{L}_{qa}$, achieves an average accuracy of 75.80\%. Consecutively adding the contrastive loss $\mathcal{L}_{c}$ and the preference losses $\mathcal{L}_{a}^{p}$ and $\mathcal{L}_{v}^{p}$ provides progressive gains. The final model, which integrates all four loss functions, achieves the highest average accuracy of 78.51\%. This demonstrates that each loss component plays a vital role, and their combined effect is crucial for optimizing the model's overall performance. It is worth noting that, as can be seen from the results in the third, fourth, and fifth rows of the table, visual preference training provides the most significant overall improvement to the model compared to other training objectives (aside from basic $\mathcal{L}_{qa}$). This also indicates that in the vast majority of scenarios, the model relies more heavily on visual information for question answering.

\begin{table}[t]
    \centering
    \caption{Different modality preference enhancements.}
    \label{tab:pe}
    \begin{adjustbox}{width=0.9\linewidth,center=\linewidth}
    {\begin{tabular}{c|c|cccc}
    \toprule
    \multirow{2}{*}{\textbf{\#}} & \multirow{2}{*}{\textbf{Methods}} & \multicolumn{4}{c}{\textbf{Average Accuracy (\%)}} \\ \cline{3-6}
    & & \textbf{A-QA} & \textbf{V-QA} & \textbf{AV-QA} & \textbf{Avg} \\
    \hline
    {1} & {w/o. APE}      & 79.21                        & 86.50                 & 73.31                    & 77.85                                   \\
    {2} & {w/o. VPE}      & 79.83                                      & 84.02                 & 73.51                    & 77.41    \\
    {3} & {w/o. AVPE}     & 77.65                                     & 85.59                  & 72.98                    & 77.15    \\
    {4} & {\textbf{AV-Master}}     & \textbf{79.95}                                     & \textbf{86.58}                  & \textbf{74.22}      & \textbf{78.51}    \\

                                     \bottomrule
    \end{tabular}}
    \end{adjustbox}
\end{table}

\begin{table}[t]
    \centering
    \caption{Different enhancement method.}
    \label{tab:add}
    \begin{adjustbox}{width=0.9\linewidth,center=\linewidth}
    {\begin{tabular}{c|c|cccc}
    \toprule
    \multirow{2}{*}{\textbf{\#}} & \multirow{2}{*}{\textbf{Methods}} & \multicolumn{4}{c}{\textbf{Average Accuracy (\%)}} \\ \cline{3-6}
    & & \textbf{A-QA} & \textbf{V-QA} & \textbf{AV-QA} & \textbf{Avg} \\
    \hline
    {1} & {W-ADD}           & 78.71                             & 81.79                  & 71.86                    & 75.70                                \\
    {2} & {MUL}         & 79.70                                     & 85.30                  & 73.70                    & 77.84                  \\
    {3} & {ADD}          & 79.95                                     & 86.58                  & 74.22                    & 78.51                  \\
                                     \bottomrule
    \end{tabular}}
    \end{adjustbox}
\end{table}

\begin{table}[t]
    \centering
    \caption{Comparison of different temporal segment selection strategies.}
    \label{tab:temporal_segment_selection}
    \begin{adjustbox}{width=0.9\linewidth,center=\linewidth}
    {\begin{tabular}{c|c|cccc}
    \toprule
    \multirow{2}{*}{\textbf{\#}} & \multirow{2}{*}{\textbf{Strategy}} & \multicolumn{4}{c}{\textbf{Average Accuracy (\%)}} \\ \cline{3-6}
    & & \textbf{A-QA} & \textbf{V-QA} & \textbf{AV-QA} & \textbf{Avg} \\
    \hline
    {1} & {Uniform}         &  77.84 & 84.92 & 72.81 & 76.91 \\
    {2} & {Top-$K$}         &  78.54 & 85.61 & 73.35 & 77.58\\
    {3} & {Gaussian}        &  78.21 & 85.23 & 73.06 & 77.24\\
    {4} & {Gaussian-Expert} & 79.07 & 86.04 & 73.71 & 78.03 \\
    {5} & \textbf{{TDPP (Ours)}}     & \textbf{79.95} & \textbf{86.58} & \textbf{74.22} & \textbf{78.51} \\
    \bottomrule
    \end{tabular}}
    \end{adjustbox}
\end{table}

\begin{figure}[t]
    \centering
    \includegraphics[width=\linewidth, keepaspectratio]{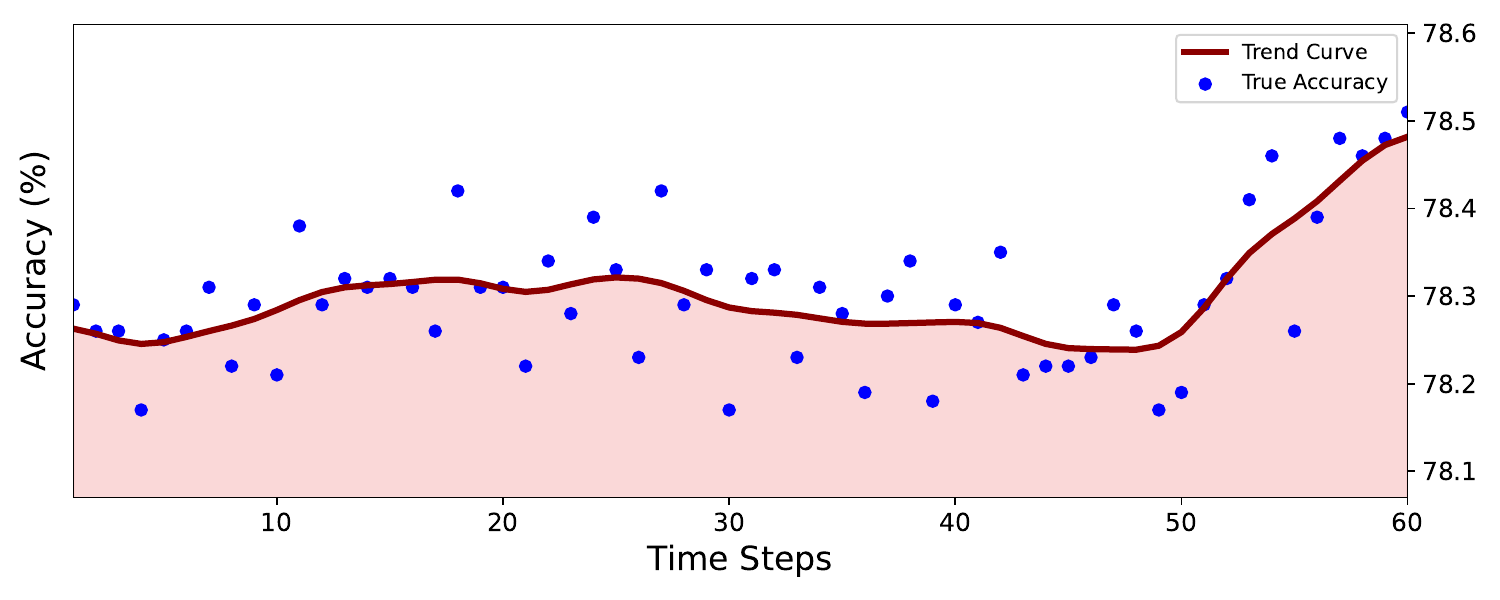}
    \caption{Intermediate answer accuracy evaluated using the focus features obtained after different temporal focus-capture steps. Blue dots denote raw accuracy at each step, and the red line denotes the smoothed trend.}
    \label{fig:4}
\end{figure}

\textit{(e)~Impact of different feature extractors:} The results in Table~\ref{tab:encoder} demonstrate that different visual encoders have a significant impact on the performance of AV-Master. When using ResNet-18~\cite{he2016deep} as the visual encoder, AV-Master achieves an average accuracy of 74.04\%, showing a considerable improvement compared to ST-AVQA~(71.52\%) with the same visual encoder. When a more powerful visual encoder, CLIP$_{\text{B}}$~\cite{radford2021learning}, is adopted, the average accuracy of AV-Master further increases to 76.31\%, surpassing PSTP-Net~(73.52\%). Moreover, when using CLIP$_{\text{L}}$~\cite{radford2021learning} and Internvideo2~\cite{wang2024internvideo2} as visual encoders, AV-Master achieved the best results across all subtasks (A-QA, V-QA, AV-QA), significantly outperforming comparable methods. In summary, as the visual encoder was progressively upgraded from ResNet-18 to Internvideo2, the performance of AV-Master on all tasks steadily improved.

Furthermore, the choice of audio encoders also plays a crucial role. When fixing the visual encoder to CLIP$_{\text{L}}$, switching from VGGish~\cite{hershey2017cnn} to a more advanced audio encoder like CLAP~\cite{wu2023large} boosts the average accuracy from 78.51\% to 79.34\%, outperforming its competitor PSOT~(78.42\%). This trend continues with the most powerful backbones. When using Internvideo2 for both visual and audio encoding, AV-Master achieves the highest overall average accuracy of 80.15\%, again surpassing the competing method CoQo~(79.60\%). These results collectively demonstrate that the performance of AV-Master is consistently enhanced by leveraging stronger feature extractors for both the visual and audio modalities, underscoring the importance of high-quality unimodal representations for complex audio-visual reasoning tasks.

\textit{(f)~Impact of different audio-visual segment lengths:} To investigate the effect of audio-visual segment lengths on model performance and the overall robustness of AV-Master in scenarios with limited audio-visual input, we conducted an ablation study on the segment length. As shown in Fig.~\ref{fig:9}, when the amount of audio–visual input decreases, the overall performance of all models shows a downward trend, indicating that more complete audio-visual information can provide richer cues and is beneficial for improving model performance. Meanwhile, compared with QA-TIGER, AV-Master maintains a relatively stable performance decline when facing reduced audio-visual content, suggesting that it is capable of extracting and integrating key features from limited audio-visual information. This can be attributed to the proposed audio–visual focus capture module, which progressively refines coarse-grained audio–visual features into fine-grained cues. In addition, the involvement of the preference activation strategy and the dual-path model architecture further strengthen the model’s robustness. Note that when the segment length is 48, AV-Master's performance shows a slight drop. This may reflect the model reaching near saturation during the middle-to-late period or being less sensitive to redundant tail segments.

\begin{figure}[!t]
    \centering
    \includegraphics[width=\linewidth, keepaspectratio]{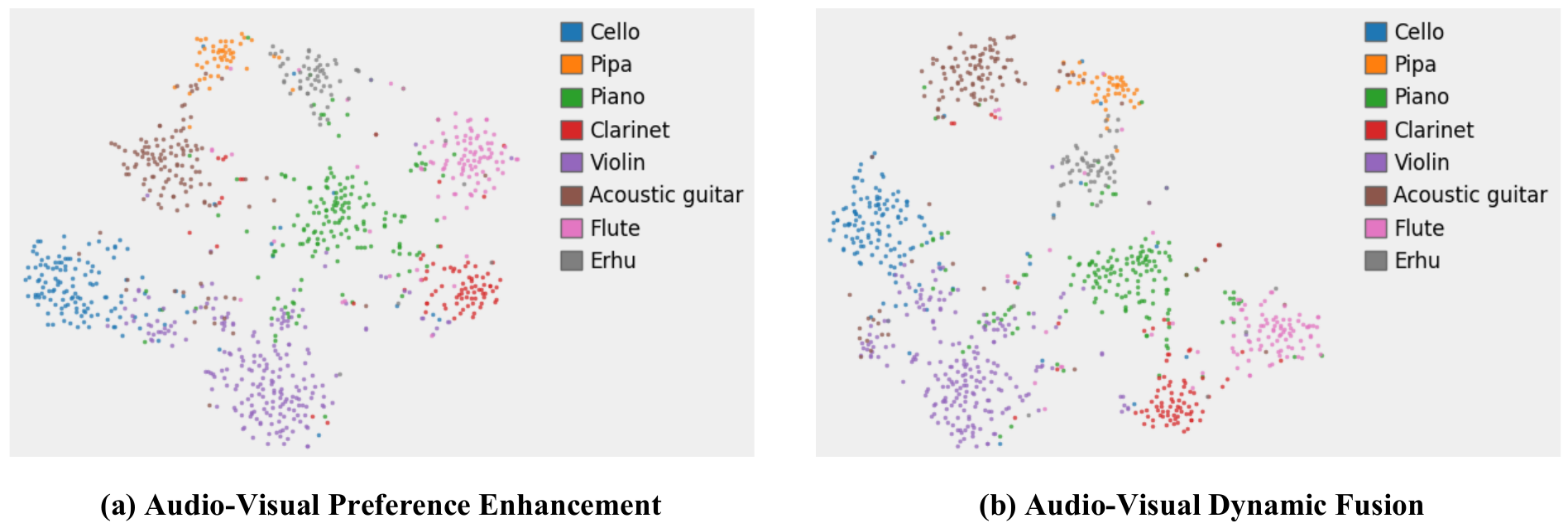}
    \caption{Comparison of t-SNE results: Preference Enhancement vs. Dynamic Fusion on Music-AVQA dataset.}
    \label{fig:tsne}
\end{figure}

\begin{figure}[!t]
    \centering
    \includegraphics[width=\linewidth, keepaspectratio]{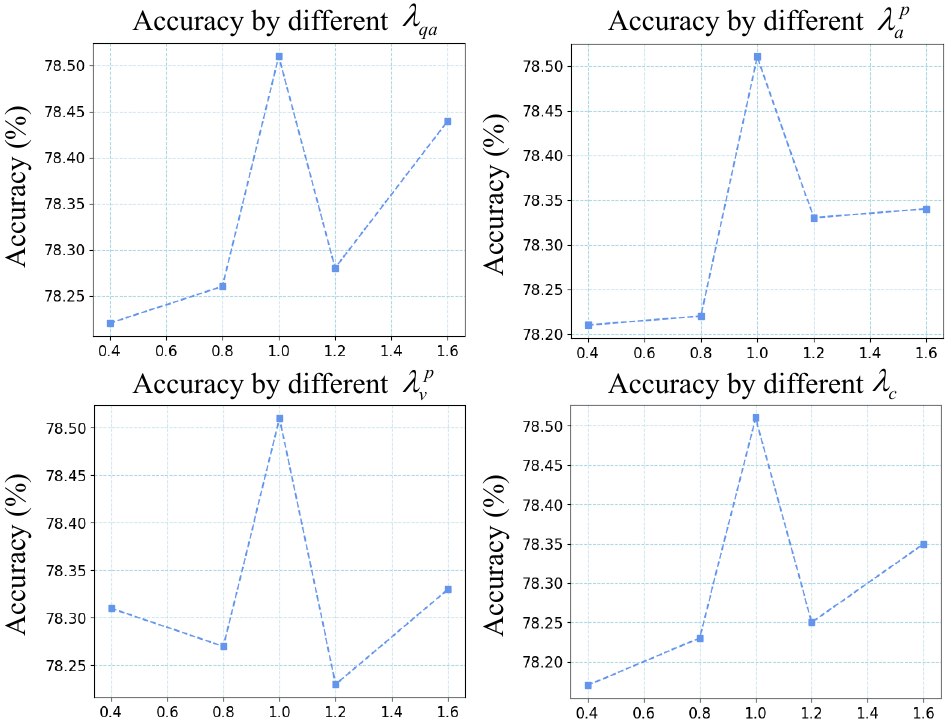}
    \caption{The effects of four trade-off parameters on the MUSIC-AVQA dataset including $\lambda_{qa}$, $\lambda_{a}^{p}$, $\lambda_{v}^{p}$ and $\lambda_{c}$.}
    \label{fig:7}
\end{figure}

\begin{figure}[!t]
    \centering
    \includegraphics[width=\linewidth, keepaspectratio]{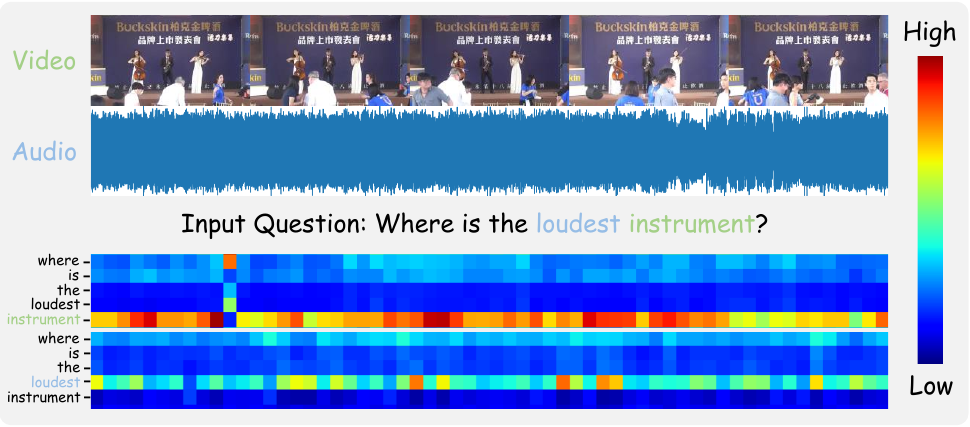}
    \caption{The attention visualization for video-question~(upper) and audio-question~(lower), with attention intensity indicated by the color scale on the right.}
    \label{fig:5}
\end{figure}

\begin{figure*}[t]
\centering
\includegraphics[width=\textwidth]{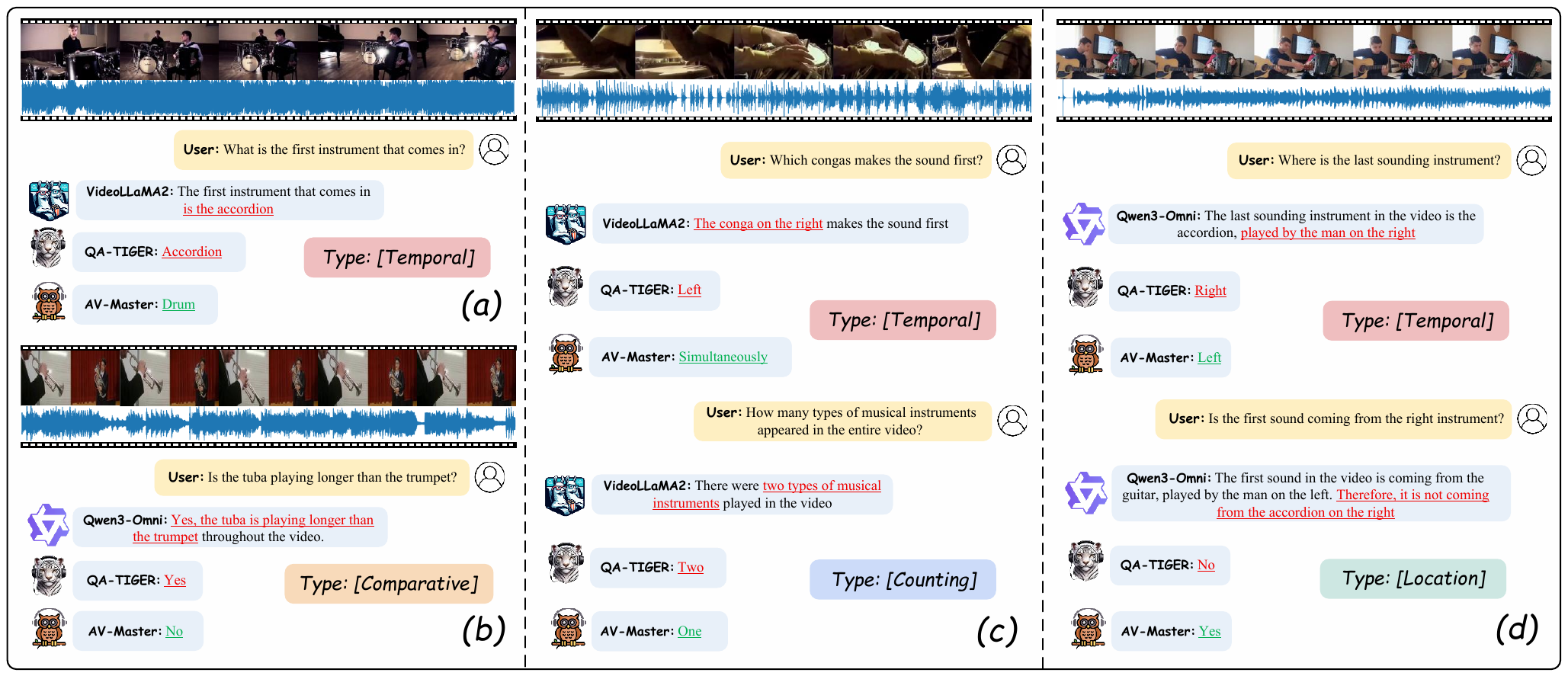} 
\caption{Qualitative demonstration of our proposed AV-Master and comparison with MLLM~(VideoLLaMa2-7B and Qwen3-Omni) and AVQA expert model~(QA-TIGER). Correct parts are highlighted in green while incorrect parts are highlighted in red.}
\label{fig:avllm}
\end{figure*}

\textit{(g)~Ablation study on input modalities:} To investigate the contribution of different input modalities and to verify the stability of the AVQA model, we conduct an ablation study using two input settings: without visual modal~(A+Q) and without audio modal~(V+Q). As shown in Fig.~\ref{fig:6}, all models demonstrate improved performance when the visual modality is present. Specifically, AV-Master achieves the highest average accuracy across both input settings, reaching 69.8\% with A+Q and 76.9\% with V+Q. Compared to the baseline model~(QA-TIGER), AV-Master shows consistent improvements across all sub-tasks~(A-QA, V-QA, AV-QA), suggesting its superior ability to extract and fuse relevant features from the input. Notably, the performance gap between A+Q and V+Q settings is larger for QA-TIGER and AV-Master than for AVST, indicating that advanced models are more effective at leveraging visual cues. These results highlight the dominant role of the visual modality in multimodal question answering, while also confirming the robustness of AV-Master under varying modality conditions. The results for QA-TIGER in \textit{(f)} and \textit{(g)} were reproduced using the official code.


\textit{(h)~Ablation study on modality preference enhancement:} To further investigate the effectiveness of our proposed modality preference enhancement strategy and to identify which preference component contributes most to the final prediction, we conduct an ablation study by selectively disabling different enhancement settings during inference. APE, VPE, and AVPE denote Audio Preference Enhancement, Visual Preference Enhancement, and Audio-Visual Preference Enhancement, respectively. Specifically, APE and VPE refer to enhancing the final answer prediction by applying the audio and visual preference distributions during inference, respectively, while AVPE denotes their joint audio-visual preference enhancement effect.  As shown in Table~\ref{tab:pe}, removing the audio-visual preference enhancement~(AVPE) leads to the most significant performance drop, reducing the average accuracy to 77.15\%, which highlights the importance of jointly modeling audio-visual scenarios. Additionally, disabling visual preference enhancement~(VPE) results in lower V-QA accuracy~(84.02\% vs. 86.58\%) and overall average accuracy~(77.41\%), indicating that visual preference plays a critical role in AVQA. Similarly, removing the audio preference enhancement~(APE) leads to degraded performance on A-QA and AV-QA, showing the necessity of modeling audio preference. Overall, the full model achieves the best performance across all metrics, confirming the complementary benefits of modality-specific preference enhancement and their joint contribution to robust AVQA performance.

\textit{(i)~Impact of different enhancement methods:} To explore the impact of different preference distribution fusion methods on model performance, we conduct a comparative study using three representative methods: (1)~W-ADD, which first computes a weighted ratio based on the initial overall scores of the visual and auditory preference distributions before performing a weighted summation. (2)~MUL, which applies element-wise multiplication between modality preferences. (3)~ADD, a simple summation of the two distributions. As shown in Table~\ref{tab:add}, the ADD strategy achieves the best overall performance, with an average accuracy of 78.51\%, outperforming both MUL~(77.84\%) and W-ADD~(75.70\%). This result suggests that a direct and unweighted summation of visual and auditory preference distributions can better preserve cross-modal complementary information and prevent overfitting to any single modality. Interestingly, although the W-ADD method introduces an adaptive weighting mechanism, its performance lags behind due to potential imbalances introduced by dynamic scaling. When viewed together with the previous ablation results in Table~\ref{tab:pe}, it is evident that both the design of modality-specific preference enhancements and the fusion mechanism of these preferences play a crucial role in optimizing overall AVQA performance.

\textit{(j)~Effectiveness of temporal segment selection strategy:} To further validate the effectiveness of the proposed Temporal Dynamic Perception Path (TDPP), we conducted additional ablation experiments by comparing TDPP with several representative temporal segment selection strategies. Specifically, we replaced TDPP with uniform sampling, Top-K selection~(from TSPM~\cite{li2024boosting}), Gaussian sampling, and Gaussian expert-based sampling~(from QA-TIGER~\cite{kim2025question}), respectively, while keeping all other network components and training settings unchanged.

As shown in Table~\ref{tab:temporal_segment_selection}, the proposed TDPP achieves the best performance among all compared strategies, demonstrating its ability to aggregate more effective temporal evidence for answer prediction. Uniform sampling treats all temporal segments equally, which may introduce redundant or less informative audio-visual evidence. Top-K selection improves performance by selecting question-relevant frames, but it may disrupt temporal continuity and increase the risk of missing transient yet critical audio-visual cues. In addition, fixed or expert-guided Gaussian sampling strategies are less flexible in modeling diverse temporal evidence patterns associated with different questions. In contrast, TDPP dynamically updates audio-visual templates and progressively perceives effective information from different temporal segments, enabling it to adaptively aggregate more discriminative audio-visual evidence while preserving temporal contextual continuity.


\textit{(k)~Impact of different trade-off hyperparameters:} As shown in Fig.~\ref{fig:7}, we investigate the impact of four trade-off hyperparameters $\lambda _{qa}$, $\lambda _{a}^{p}$, $\lambda _{v}^{p}$ and $\lambda _{c}$, which are used to balance different loss functions. To assess the importance of each loss function in AV-Master, we vary one hyperparameter at a time while keeping the others fixed at a default value of 1.0. The experimental results show that increasing $\lambda _{qa}$ from 0.4 to 1.0 significantly improves accuracy. However, when the value increases beyond a certain threshold~(e.g., 1.6), accuracy declines slightly, suggesting that overemphasizing a single objective may diminish the contributions of others. Similar patterns are observed for $\lambda _{a}^{p}$, $\lambda _{v}^{p}$ and $\lambda _{c}$, underscoring the independent yet equally important roles of each objective in the AV-Master. Based on these findings, we set all trade-off parameters to a default value of 1.0 to ensure a balanced contribution from each loss component. This configuration achieves optimal overall performance.


\subsection{Qualitative Analysis}


\textit{(a)~Focus effects at different time steps:} As shown in Fig.~\ref{fig:4}, we present the variation in accuracy over time during the audio-visual focus capture process. The blue dots in the figure represent the original accuracy data at different moments, while the red curve indicates the overall trend. Although the original data exhibits some fluctuations, the trend line shows that the accuracy remains relatively stable during the first 40 steps. However, starting at step 45, the accuracy begins to rise significantly, reaching its peak at step 60. This suggests that the audio-visual focusing mechanism becomes more effective in the later stages, and the model progressively improves its ability to extract and fuse relevant information during the focusing process. 

\textit{(b) t-SNE comparison between preference enhancement and dynamic fusion}: As shown in Fig. 7, we provide a t-SNE visualization on the MUSIC-AVQA dataset to compare the learned representations of the proposed audio-visual preference enhancement strategy and the Dynamic Fus variant. It can be observed that the features produced by our preference enhancement strategy exhibit more compact intra-class clusters and clearer inter-class boundaries. In particular, several instrument categories show more concentrated distributions with less overlap with neighboring classes, indicating stronger discriminative capability of the learned representations. In contrast, the Dynamic Fus variant presents relatively more scattered feature distributions and more pronounced overlap among some categories, suggesting that estimating modality importance only after multimodal fusion is more susceptible to interference from entangled cross-modal information. This can be attributed to the fact that our method, under question guidance, directly models modality preference from the initial audio and visual features, rather than inferring modality weights only at the fusion stage. Such a design activates question-relevant global information while preserving modality-specific semantics, enabling the model to learn a more structured and separable feature space.

\textit{(c)~Attention visualization of different modalities:} As shown in Fig.~\ref{fig:5}, we illustrate which parts of the text the visual and audio modalities attend to within the global preference activation path. For the input question “Where is the loudest instrument?”, the visual modality primarily attends to the word “instrument” while paying minimal attention to the audio-related word “loudest.” In contrast, the audio modality places significant focus on “loudest.” These results indicate that AV-Master can effectively distinguish between visual and auditory cues and accurately align them with the corresponding textual elements, which further demonstrates AV-Master’s capability in cross-modal understanding.

\begin{figure*}[t]
\centering
\includegraphics[width=\textwidth]{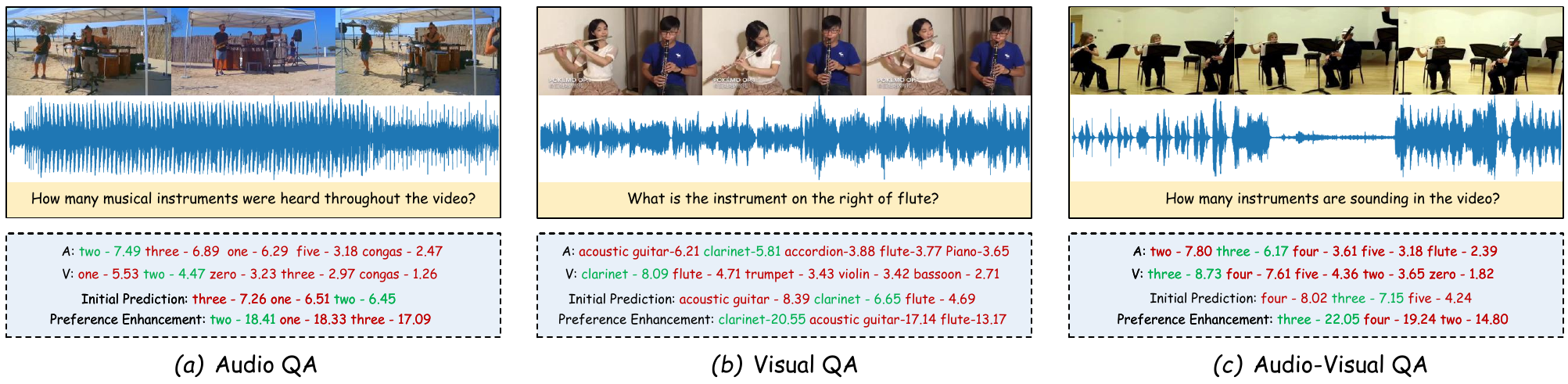} 
\caption{Visualization of the proposed preference enhancement strategy on Audio QA, Visual QA, and Audio-Visual QA examples. A and V denote the audio and visual branches in GPAP, respectively. Initial Prediction denotes the direct prediction without preference enhancement, while Preference Enhancement denotes the final prediction after incorporating audio-visual preference distributions. Correct answers are highlighted in green and incorrect answers in red.}
\label{fig:pe}
\end{figure*}

\textit{(d)~Qualitative results comparison of different models:} Fig.~\ref{fig:avllm} presents a qualitative comparison among our proposed AV-Master, MLLM~(VideoLLaMA2~\cite{damonlpsg2024videollama2} and Qwen3-Omni~\cite{yang2025qwen3}), and another AVQA expert model~(QA-TIGER~\cite{kim2025question}) across four distinct audio-visual scenarios. In each case, AV-Master exhibits a superior ability to comprehend audio-visual content and provides accurate answers that align closely with both visual and auditory cues. In contrast, the other models fail to achieve such consistency. For example, in the first scenario~(a), both VideoLLaMA2 and QA-TIGER incorrectly identify the instrument as an ``accordion'', whereas AV-Master correctly identifies it as a ``drum'', owing to its accurate interpretation of the audio signal. Similarly, in the other scenarios~(c-d), AV-Master accurately infers the spatial and temporal characteristics of sound events, while the competing models fail to effectively integrate these multimodal cues. These qualitative results reinforce our quantitative findings, showing that AV-Master is more robust in fine-grained audio-visual reasoning, particularly in scenarios that require synchronized and cross-modal understanding. This further validates the effectiveness of our carefully designed modules in fully leveraging both audio and visual modality information. The inference results for VideoLLaMA2 and Qwen3-Omni were obtained from its official demo, simulating application scenarios in low-resource environments.

\textit{(e)~Visualization of preference enhancement:} To further analyze how the proposed preference enhancement strategy affects the final decision, we visualize the top-5 logits of the audio and visual preference branches in GPAP, as well as the top-3 logits before and after preference enhancement. As shown in Fig.~\ref{fig:pe}, green denotes the ground-truth and red denotes incorrect candidates. Here, A and V represent the audio and visual branches in GPAP, respectively. Initial Prediction denotes the direct prediction without preference enhancement, while Preference Enhancement denotes the final prediction after incorporating the audio-visual preference distributions.

In the Audio QA example, the question asks how many musical instruments are heard throughout the video. The audio branch assigns the highest logit to the correct answer ``two''. Although the initial prediction incorrectly selects ``three'', preference enhancement substantially increases the logit of ``two'' and corrects the final answer. This demonstrates that GPAP can effectively activate audio-dominant evidence for sound-centric questions. In the Visual QA example, the model needs to identify the instrument on the right of the flute. The visual branch clearly highlights the correct answer ``clarinet''. After preference enhancement, the prediction is corrected from ``acoustic guitar'' to ``clarinet'', indicating that the proposed strategy can emphasize the more informative modality according to the question semantics.

In the Audio-Visual QA example, the question requires counting the instruments that are sounding in the video, where both auditory and visual evidence are necessary. The initial prediction incorrectly favors ``four''. After preference enhancement, the correct answer ``three'' receives the highest logit by integrating complementary audio and visual preferences. These qualitative results show that the proposed preference enhancement strategy not only strengthens modality-specific evidence for audio-only or visual-only questions, but also improves joint audio-visual reasoning by suppressing misleading candidates and promoting the ground-truth answer.


\section{Conclusion}\label{sec:5}

In this paper, we propose AV-Master, a novel dual-path audio-visual question answering expert model designed to address the challenges faced by existing models in processing complex audio-visual scenes. Current methods often struggle to flexibly focus on the most question-relevant spatiotemporal segments when confronted with a large amount of redundant information, and they lack the ability to dynamically perceive the importance of different modalities for different questions. This limits their comprehensive understanding of audio-visual scenes. To tackle these difficulties, AV-Master introduces a dynamic adaptive focus sampling mechanism and a global modality preference activation strategy. These enable it to effectively capture question-relevant audio-visual segments and modality preferences, thereby enhancing its decision-making capabilities. Extensive experiments on multiple large-scale AVQA benchmark demonstrate that by meticulously capturing key audio-visual details and integrating a global understanding of modality preferences, AV-Master provides an efficient and powerful solution for AVQA, particularly excelling in complex reasoning tasks and under challenging data distributions.

\bibliographystyle{ieeetr}
\bibliography{New_IEEEtran_how-to}


\end{document}